\RequirePackage{fix-cm}
\documentclass[smallextended]{svjour3}       % onecolumn (second format)
\smartqed  % flush right qed marks, e.g. at end of proof
\usepackage{cite}
\usepackage{amsmath,amssymb,amsfonts,bm}
\usepackage{algorithm}
\usepackage{algorithmic}
\usepackage{subfigure}
\usepackage{graphicx}
\usepackage{epstopdf}
\usepackage{float}
\usepackage{caption}
\usepackage{textcomp}
\usepackage{xcolor}
\usepackage{authblk}
\usepackage[marginal]{footmisc}
\usepackage{multirow}
\usepackage{multicol}
\usepackage{geometry}
\usepackage[figuresright]{rotating}
\begin{document}

\title{Multimodal-Aware Weakly Supervised Metric Learning with Self-weighting Triplet Loss
}


\author{Huiyuan Deng         \and
        Xiangzhu Meng        \and
        Lin Feng$^\ast$ \thanks{Corresponding author (denoted by $\ast$): Lin Feng}
}


\institute{H. Deng \at
              School of Computer Science and Technology, Dalian University of Technology, Dalian, China \\
              \email{dhytorres@mail.dlut.edu.cn}           %  \\
           \and
          X. Meng \at
              School of Computer Science and Technology, Dalian University of Technology, Dalian, China \\
              \email{xiangzhu\_meng@mail.dlut.edu.cn}
          \and
          L. Feng \at
              School of Computer Science and Technology, Dalian University of Technology, Dalian, China \\
              \email{fenglin@dlut.edu.cn}
}

\date{Received: date / Accepted: date}

\maketitle

\begin{abstract}
In recent years, we have witnessed a surge of interests in learning a suitable distance metric from weakly supervised data. Most existing methods aim to  pull all the similar samples closer while push the dissimilar ones as far as possible. However, when some classes of the dataset exhibit multimodal distribution, these goals conflict and thus can hardly be concurrently satisfied. Additionally, to ensure a valid metric, many methods require a repeated eigenvalue decomposition process, which is expensive and numerically unstable. Therefore, how to learn an appropriate distance metric from weakly supervised data remains an open but challenging problem. To address this issue, in this paper, we propose a novel weakly supervised metric learning algorithm, named MultimoDal Aware weakly supervised Metric Learning (MDaML). MDaML partitions the data space into several clusters and allocates the local cluster centers and weight for each sample. Then, combining it with the weighted triplet loss can further enhance the local separability, which encourages the local dissimilar samples to keep a large distance from the local similar samples. Meanwhile, MDaML casts the metric learning problem into an unconstrained optimization on the SPD manifold, which can be efficiently solved by Riemannian Conjugate Gradient Descent (RCGD). Extensive experiments conducted on 13 datasets validate the superiority of the proposed MDaML.
\keywords{Weakly Supervised Metric Learning \and Riemannian Metric \and Similarity Measures}
\end{abstract}

\section{Introduction}
With the development of the information era, multimodal distribution data \cite{64, 65, 66} frequently exist in real-world applications, for example, the distribution of medial checkup samples of sick patients \cite{55} and the distribution of the wind \cite{66} are kinds of the common multimodal distribution data. Multimodal distribution phenomenon occurs when data instances sharing the same class label forming several separate clusters \cite{67}. Measuring the similarities between samples is the fundamental task in machine learning area \cite{68,69, 63, 71, 79, 80}.
However, their performance has a severely drop when facing with the multimodal distribution dataset \cite{50, 64}.
Therefore, how to measure similarities between samples from multimodal distribution data is of vital importance.
\par
Recently, weakly supervised distance metric learning (DML), typically referring to learning a Mahalanobis distance metric from the side information has received considerable attention in various data mining tasks due to its wide applications \cite{16,18,19,20,70, 76,77, 78}. The side information is provided with the form of pairwise or triplet constraints.
In general, weakly supervised DML methods aim to learn a metric such that the distances between the semantic similar samples are smaller than those between examples from semantic dissimilar ones \cite{1, 5, 21, 8, 9, 21, 35}. For example, in \cite{1}, the author proposes to learn an appropriate metric by minimizing the sum of the distances between the similar pairs while enforcing the sum of the square distances between the dissimilar pairs larger than a predefined threshold. The metric learning can be formulated in a convex optimization framework, which can be solved by projected gradient descent \cite{5,28}.
\par
\begin{figure*}[htbp]
  \centering
  \subfigure[Before Learning]{
  \label{before_learning}
 \includegraphics[width = 0.3\textwidth]{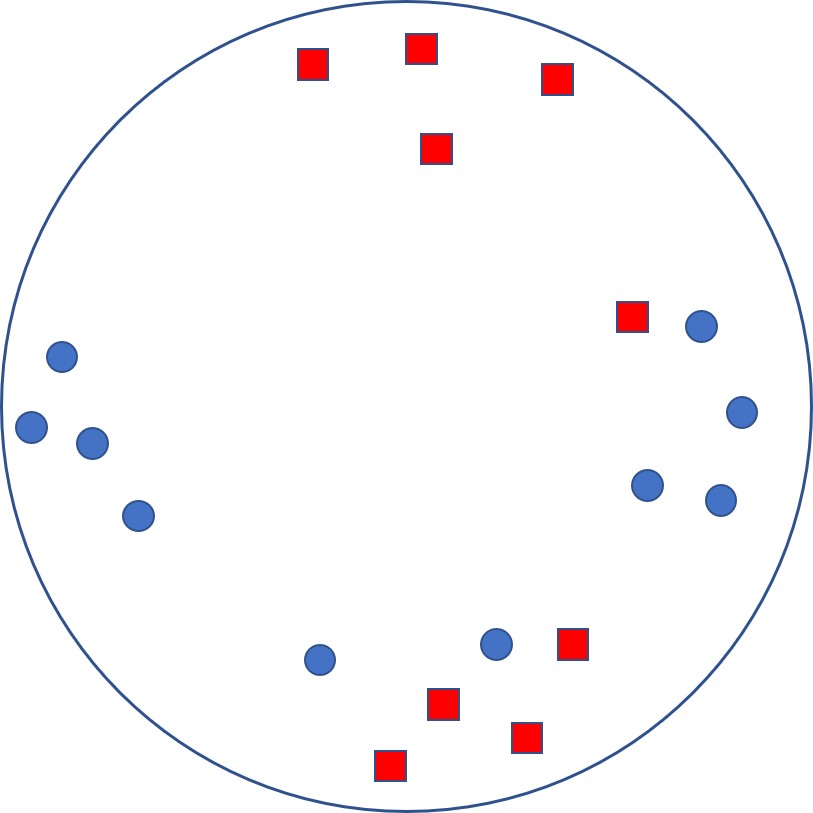}}
  \subfigure[During Learning]{
\label{during learning}
 \includegraphics[width = 0.3\textwidth]{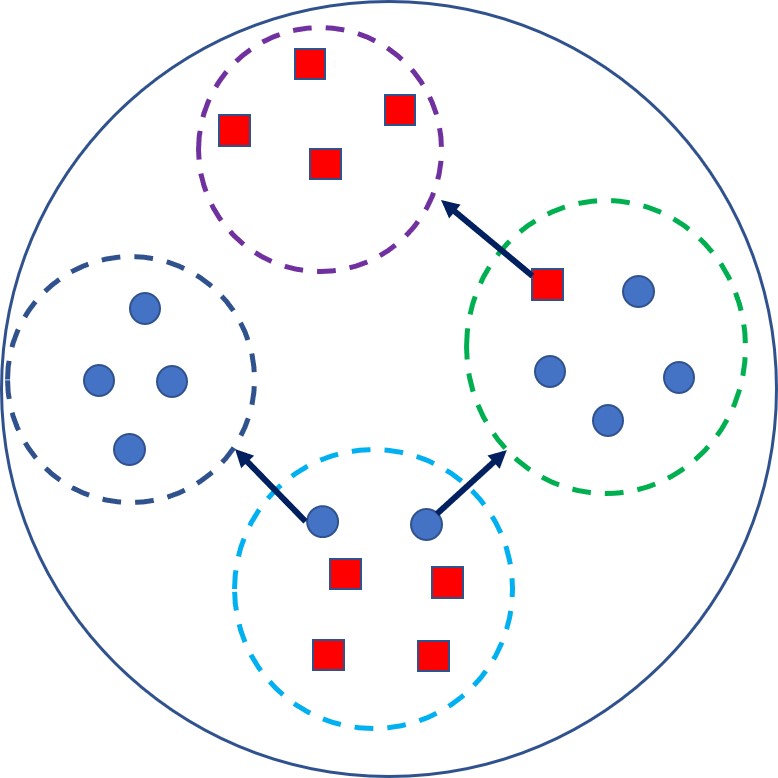}}
  \caption{The schematic illustration of MDaML (a) before learning; (b) during learning. The blue circle and red square represent two different classes. MDaML first partitions the data space into some clusters based on the classes or the different modes, then, it optimizes the local compactness and separability.}
  \label{illustration pictures}
\end{figure*}

Despite the promising efforts that have been made. most existing weakly supervised DML methods are confronted in that they treat all the side information equally important and aim to learn a metric to satisfy all the constraints. Such paradigm based methods may suffer from sub-optimal learning performance when facing some real-world datasets with multimodal distribution. Take the Figure .\ref{before_learning} as an example, the red square and blue circle are two classes of instances that lie on the two-dimensional feature space, enforcing the sample classes instances closer to each other only resulting in a mixture of the two classes.
One common way to alleviate this limitation is to identify those local constraints and optimize the local compactness and separability. Local Distance Metric (LDM) \cite{62} utilizes the kernel function to identify the local similar pairs and proposes a probabilistic framework to optimize the leave-one-out evaluation using the $k$NN classifier. In \cite{72}, a local variant of the ITML \cite{21} method is proposed to design an adaptively local decision rule to relax the fixed threshold. These two works both leverage the information from the local constraints to guide the DML process.
However, they fail to fully exploit the underlying data distribution to distinguish the local side information.
\par
\par
In this paper, we propose a novel weakly supervised DML method named Multimodal Distribution-aware weakly supervised Metric Learning (MDaML). It can effectively exploit underlying data distribution to discover the different modes for multimodal data and decreases the distance of similar pair from the same mode while enlarges the distance of dissimilar pair from the same mode. More specifically, DMaML formulates the objective function as the combination of the clustering term and the sum of the weighted triplet loss term. The clustering term is utilized to partition the whole train data into several clusters based on their mode, and the weighted triplet loss is added to enhance the local separability. Different from most existing DML methods depending on the repeated eigenvalue decomposition process, our proposed method cast the ML problem into an unconstrained optimization on the SPD manifold, which can be efficiently solved by Riemannian Conjugate Gradient Descent (RCGD) method.

\subsection{Contributions}
The major contributions in this paper can be summarized as follows:
\begin{itemize}
    \item {A novel weakly supervised DML method is proposed to handle the multimodal distribution data, which explicitly separates the different distributions and optimizes the local separability in each mode.}
    \item{To solve MDaML, we develop an effective method based on iterative alternating strategy meanwhile casting the DML problem into an unconstrained optimization on the SPD manifold.}
    \item{Experiments results on various datasets validate that our proposed algorithm has a reasonable training time and can improve the performance on the classification task.}
\end{itemize}
\par
\begin{table*}[htbp]
\centering
\caption{Important notations.}
\label{notation}
\begin{tabular*}{\textwidth}{@{\extracolsep{\fill}}ll}  % {lccc} left-l,right-r,center-c
\hline
Notation &  Description  \\
\hline  % \hline
$K$ & The number of localities centers.\\
$d$ & The dimensionality of the data.\\
$N$ & The number of instances in the training set.\\
$\bm{x}_{i}$ &  The $i$-th sample of the training set.\\
$\mathbf{X} = \left\{ \bm{x}_{i} \vert \bm{x}_{i} \in \mathbb{R}^{d} \right \}$ & The raw data matrix.\\
$\mathcal{T} = \{(\bm{x}_{i}^t, \bm{x}_{j}^t, \bm{x}_{r}^t)\}_{t = 1}^{T}$ & The triplet side information. \\
$\mathcal{S}_{d}^{+}$ & The positive semi-definite (PSD) cone.\\

$dis_{M}(\bm{x}_{i}, \bm{x}_{j})$ & The Mahalanobis distance between $\bm{x_i}$ and $\bm{x_{j}}$ under $\mathbf{M}$.\\
$\mathcal{W} = \left \{\bm{w}_{i} \in \mathbb{R}^{K} \right \}_{i = 1}^N$ & The weight set of the training set.\\
$\left |\left | \mathbf{M} \right| \right|_F^{2}$ & The square Frobenius Norm of the matrix $\mathbf{M}$.\\
$\mathcal{C}_K = \left \{\bm{c}_{k} \in \mathbb{R}^{N} \right \}_{k = 1}^K$ &  The set of the locality centers. \\
$\mathcal{S}_{++}^{d}$ & The set of real symmetric positive definite matrix (SPD).\\

\hline
\end{tabular*}

\end{table*}

\subsection{Organization}
The rest of the paper is organized as follows: in Section \uppercase\expandafter{\romannumeral2}, we briefly review some related work to the proposed method; in Section \uppercase\expandafter{\romannumeral3}, we describe the construction procedure of MDaML and optimization algorithm for MDaML in detail; in Section \uppercase\expandafter{\romannumeral4}, empirical evaluations based on the applications of benchmark classification and image classification demonstrate the effectiveness of our proposed approach; in Section \uppercase\expandafter{\romannumeral5}, we make a conclusion of this paper.

\section{Related Works}
During the last decades, metric learning has received considerable attention due to its excellent performance in measuring the similarities between samples. Here, we confine ourselves to the discussion of several classical or related DML works. For a thorough study of metric learning, please refer to \cite{32, 45}.
The MMC \cite{1} is considered to be the pioneering work of DML, this method aims to minimize the sum of the distance between similar pairs $\mathcal{S}$ while keeps the sum of the square distance between dissimilar pair $\mathcal{D}$ larger than a predefined threshold. Logistic discriminant metric learning (LDML) \cite{7} proposes a logistic discriminant approach to obtain a metric using the pairwise constraints. Information-theoretic metric learning (ITML) \cite{21} learns the metric by minimizing the differential relative entropy between two multivariate Gaussians, the distance constraints are further added to improve the separability of the modal. while these methods can make significant improvement, most of them neglect the existence of multimodal distribution classes in the dataset and fail to take the holistic distribution of the dataset into consideration.
\par
To tackle this drawback, some works focus on learning the local metric. Large margin nearest neighbor (LMNN) \cite{5} trains a metric with the goal that all the $k$-nearest neighbors always belong to the same class and keeps a large margin from the sample form different class. Local Fisher Discriminant Analysis (LFDA) \cite{50} is local variant FDA and takes local structure of the data into consideration. Training a local distance metric can cope with the multimodal distribution data by wisely building the side information using the local similar and dissimilar pairs. Nevertheless, all these method require the instance label to help choose the suitable local pairs. Different with them, our proposed method learns a metric in a weakly supervised setting and optimizes the local compactness and separability for each mode, therefore, MDaML have a broader application scenarios.
\par
Local Distance Metric (LDM) \cite{62} is a weakly supervised DML that aims to learn a metric by employing eigenvector analysis and bound optimization strategy in a probabilistic framework. However, it is different with our work. LDM utilize the kernel function to identify the local similar pairs, while we use a explicit clustering strategy to cluster the data into several clusters according to their modes. Our work shares the same spirit but with a totally different formulation.
\par
Multi-metric learning methods aim to learn multiple distance metrics according to the different multiple latent semantics\cite{61} or the different spatial localities \cite{73, 5}, and the final distance of the two instances is the combination of the weighted component metrics. These methods can deal with the multimodal distribution data well. However, these methods are harder to optimize and are more prone to over-fitting sometimes\cite{74}.

\section{Preliminaries}
In this section, some basic knowledge of DML are introduced first, then we review the geometry and operations on SPD manifold. For convenience, some important notations used in this paper are listed in TABLE \ref{notation}.

\subsection{DML Problem Definition}
Suppose we are given a dataset containing $N$ instances, $\mathbf{X} \in \mathbb{R}^{d \times N}$, instance $\bm{x}_{i} \in \mathbb{R}^{d}$ is sampled from a $d$-dimensional unknown feature space. The weakly supervised information are provided in the form of pairwise or triplet constraints. The pairwise constraints sets consists of the semantic similar pair set $\mathcal{S}$ and semantic dissimilar pair set $\mathcal{D}$.\\
\begin{equation*}
\begin{split}
    &\mathcal{S}:= \left\{(\bm{x}_{i},\bm{x}_{j})\  \vert  \ \bm{x}_{i}\ \text{and}\  \bm{x}_{j} \ \text{are semantic similar} \right \}\\
   &\mathcal{D}:= \left\{(\bm{x}_{i},\bm{x}_{j})\  \vert \  \bm{x}_{i}\ \text{and}\  \bm{x}_{j} \ \text{are semantic dissimilar} \right \}.
\end{split}
\end{equation*}
Another widely used side information is the triplet constraints $\mathcal{T}$ which is made up with three elements.
\begin{equation*}
\begin{split}
\mathcal{T}:= &\left \{(\bm{x}_{i}, \bm{x}_{j}, \bm{x}_{r})\ \vert \ \bm{x}_{i} \ \text{is more similar to} \  \bm{x}_{j}\  \text{than to}\  \bm{x}_{r}\right \}.
\end{split}
\end{equation*}
For any $\bm{x}_{i}$ and $\bm{x}_{j}$, the (square) Mahalanobis distance between this two instances is calculated as:
\begin{equation}
    dis_{\mathbf{M}}^2(\bm{x}_{i}, \bm{x}_{j}) = (\bm{x}_{i} - \bm{x}_{j})^T\mathbf{M}(\bm{x}_{i} - \bm{x}_{j})
\end{equation}
where $\mathbf{M}$ is a symmetric Positive Semi-Definite (PSD) matrix. When $\mathbf{M} = \mathbf{I}$, the Mahalanobis metric degenerates to Euclidean distance metric. since Mahalanobis metric matrix takes the features correlations and weights into consideration, they can better measure the similarity between samples than the Euclidean distance metric \cite{1, 61, 75}. The goal of weakly supervised DML is to learn a Mahalanobis metric matrix from the side information, such that semantic similar data points are pulled close to each other while dissimilar data points are pushed away \cite{6, 7, 8}.

\subsection{Optimization on the SPD Manifolds}
In this section, we review the geometry and operations on SPD manifolds.
\par
{\bfseries The SPD Manifold.} The set of $(p \times p)$ dimensional real, SPD matrices endowed with the Affine Invariant Riemannian Metric(AIRM)\cite{42} forms the SPD manifold $\mathcal{S}_{++}^p$.
\begin{equation}
    \mathcal{S}_{++}^p = \{\mathbf{M} \in \mathbb{R}^{p \times p}: \mathbf{M}^T = \mathbf{M},
    \bm{v}^T\mathbf{M}\bm{v} > 0, \forall{\bm{v}} \in \mathbb{R}^p\setminus \{\bm{0}_p\} \}
\end{equation}
where $\bm{0}_p$ represents the $p$-dimensional vector whose elements are all $0$. For any point $\mathbf{W} \in \mathcal{S}_{++}^p$, there exists a tangent space $\mathcal{T}_W\mathcal{S}_{++}^p$, which is tangent to all the smooth curves within $\mathcal{S}_{++}^p$ passing through $\mathbf{W}$.
\par
{\bfseries Orthogonal projection on SPD Manifolds}. The orthogonal projection operation transforms the Euclidean gradient from the ambient space onto the tangent space. On SPD manifolds, the Orthogonal projection operation $\pi_{W}(\cdot): \mathbb{R}^{p\times p} \rightarrow \mathcal{T}_{\mathbf{W}}\mathcal{S}_{++}^p$ is defined by
\begin{equation}\label{riemann_gd}
    \pi_{\mathbf{W}}(\nabla_{\mathbf{W}}) = \mathbf{W}\frac{1}{2}({\nabla_{\mathbf{W}} + \nabla_{\mathbf{W}}^T)\ \mathbf{W}}
\end{equation}
where $\mathbf{W} \in \mathcal{S}_{++}^P$, $\nabla_{\mathbf{W}}$ is the euclidean gradient of any smooth function at point $\mathbf{W}$.
\par
{\bfseries Retraction operation On SPD Manifolds}. The retraction operation transforms the point on the tangent space to the manifold, and must satisfy centering and local rigidity properties \cite{38}. For the SPD manifolds, the idea retraction is called exponential map: $\Gamma_W(\cdot):\mathcal{T}_W\mathcal{S}_{++}^P \rightarrow \mathcal{S}_{++}^d$, given by
\begin{equation}
    \Gamma_W(\mathbf{\zeta}) = \mathbf{W}^{\frac{1}{2}}\text{expm}(\mathbf{W}^{-\frac{1}{2}} \mathbf{\zeta} \mathbf{W}^{-\frac{1}{2}})\mathbf{W}^{\frac{1}{2}}
\end{equation}
where $\mathbf{\zeta} \in \mathbb{R}^{d\times d}$ is a point on the tangent space, and the tangent space is at $\mathbf{W}$. The $\text{expm}(\cdot)$ is the matrix exponential operation.
\par
To perform the Riemannian gradient descent procedure for the loss function $\mathcal{L}$ at the point $\mathbf{W}^{t}$ on the SPD manifolds, we can utilize the "projection and retraction" strategy. First, computing the Euclidean gradient of the loss function with respect to $\mathbf{W}^{t}$, denoted as $\nabla{\mathcal{L}(\mathbf{W}^{t})}$. Then, projecting the Euclidean gradient to the tangent space $\mathbf{\zeta}^t = \pi_{\mathbf{W}^t}(\nabla{\mathcal{L}(\mathbf{W}^{t})})$. Next, calculating the new point by $\mathbf{W}_{+}^t = \mathbf{W}^{t} - \alpha \mathbf{\zeta}^t$, where $\alpha$ is the step size. Last, retracting the $\mathbf{W}_{+}^t$ back to the manifold: $\mathbf{W}^{t+1} = \Gamma_{\mathbf{W}^t}(\mathbf{\mathbf{W}_{+}^t})$. With a proper step size and enough iteration, this procedure can be proven to convergence at last \cite{38}.

\section{MDaML Approach}
In order to train an appropriate distance metric for weakly supervised multimodal distribution data, we propose a novel DML method called MDaML in this section. First, we elaborate the motivation of the proposed method in \uppercase\expandafter{\romannumeral3}-A. Then, the detail
construction of proposed method is introduced in Section \uppercase\expandafter{\romannumeral3}-B. Finally, the optimization procedure is described in Section\uppercase\expandafter{\romannumeral3}-C.

\subsection{Motivation}
Even though weakly supervised DML has attracted increasing attention during the last decade, most of existing weakly supervised DML methods fail to provide appropriate distance metric to measure similarities between samples from multimodality dataset. These algorithms devote to devising various objection functions to satisfy the pairwise or triplet constraints and fail to take the holistic data distribution into consideration. To tackle this problem, this paper intend to learn a novel distance metric which can fully exploit underlying data distribution.
\subsection{Construction process of MDaML}
The detail construction process of the proposed MDaML is illustrated in this section. For most of existing weakly supervised DML methods that aim to obtain an appropriate metric in a global sense, however, when facing with multimodal distribution data, the performance of these methods will have a severely drop due to the conflict of some side information. Therefore, it is vital importance for MDaML to give less attention to those misleading side information and train a local metric. First of all, MDaML assumes that the training data forms several clusters as a matter of the multimodal distribution or the difference of the classes. This assumption ensures that a standard clustering algorithm can partition the data space into several parts and helps us to identify the local side information. Then, the weighted triplet loss \cite{61} is added to increase the local discriminative ability. Finally, to mitigate the risk of overfitting, we combine the square frobenius norm of the metric matrix with the objective function.
\par
 We assume there are $K$ local clusters centers $\mathcal{C}_K = \{\bm{c}_k \in \mathbb{R}^d\}_{k = 1}^K$ anchoring over the whole data space, and a corresponding probabilistic weight set $\mathcal{W} = \{(w_{i1}, w_{i2}, ...,w_{iK})^T\}_{i=1}^N$ for each training data. The value of $w_{ik}$ can reflect the affinity between $\bm{x}_i$ and $\bm{c}_k$, specifically, the closer $\bm{x}_i$ is to $\bm{c}_k$, the larger value of $\bm{c}_k$, this can be achieved by minimizing a clustering-like objective function Eq. (\ref{cluster}), which enforces the instances to aggregate toward their own anchor centers induced by the metric matrix $\mathbf{M}$.\\
\begin{equation}\label{cluster}
   \begin{split}
    &\min \limits_{\mathbf{M}, \mathcal{C}_K, \mathcal{W}}\ \sum_{i = 1}^N \sum_{k =1}^K w_{ik}^{\eta}(\bm{x}_i - \bm{c}_k)^T\mathbf{M}(\bm{x}_i - \bm{c}_k)\\
    & s.t. \quad w_{ik} > 0; \ \sum_{k =1}^K w_{ik} = 1; \ \forall{1 \leq i \leq N};\  \mathbf{M} \in \mathcal{S}_{d}^{+}.
    \end{split}
\end{equation}
where $\eta > 0$ is the smooth parameter.
\par
In the literature, triplet constraints are wildly employed in the weakly supervised DML \cite{63, 61, 33, 3}. Different from pairwise constraints methods, DML methods using triplet constraints adopt the relative large margin criterion and separate the semantic dissimilar sample from the semantic similar sample by a large margin. Given the $\mathcal{T} =\left \{\bm{x}_{i}^{t},\bm{x}_{j}^{t},\bm{x}_{r}^{t}\right\}_{t=1}^{T}$ as the side information, the triplet loss can be expressed as:
 \begin{equation}\label{5}
    \begin{split}
    \min \limits_{\mathbf{M}} & \ \frac{1}{T}\sum_{t = 1}^{T} \ell(dis_{\mathbf{M}}^2(\bm{x}_{i}^{t}, \bm{x}_{r}^{t})-dis_{\mathbf{M}}^2(\bm{x}_{i}^{t}, \bm{x}_{j}^{t}))\\
    &\mathbf{M} \in \mathcal{S}_{d}^{+}.\\
    \end{split}
 \end{equation}
 where $\ell(\cdot)$ is a convex non-increasing loss function, the larger the input value, the smaller the loss. It encourages the square distance between the dissimilar pair larger than that of the similar pair.
 The triplet loss function treats every triplet constraint equally important and aims to learn a metric to satisfy all the triplet constraints. However, this goal conflict when facing with multimodal distribution data. We argue that when the similar pair in a triplet constraint from two different mode, they should not be forced to be close but to be stay aside. It is the local structure that determine the goodness of the learned matrix, specifically, the closer of the two similar point, the more degree that should match the triplet constraint, otherwise, less concerned should be focus on the triplet constraint that violates the constraints. We utilize $\mathcal{W}$ to decide the closeness of the similar pair $(\bm{x}_{i}^{t}, \bm{x}_{j}^{t})$ in a triplet $(\bm{x}_{i}^{t}, \bm{x}_{j}^{t}, \bm{x}_{r}^{t})$. According to the Eq.(\ref{cluster}), the value of $w_{ik}$ can reflect the closeness of the point $\bm{x}_{i}^{t}$ and anchor center $\bm{c}_{k}$, therefore, we combine the Eq.(\ref{cluster}) and Eq.(\ref{5}) to formulate the objective function.
 \begin{equation}\label{new1}
     \begin{split}
         &\min \limits_{\mathbf{M}, \mathcal{C}_K, \mathcal{W}}{\mathcal{L}}(\mathbf{M}, \mathcal{C}_K, \mathcal{W})\\
         &= \frac{1}{NK}\sum_{i = 1}^N \sum_{k =1}^K w_{ik}^{\eta}(\bm{x}_{i} - \bm{c}_{k} )^T\mathbf{M}(\bm{x}_{i} - \bm{c}_{k}) \\
         &+ \frac{\lambda_1}{KT}\sum_{k =1}^K \sum_{t = 1}^{T}w_{ik}^{\eta} w_{jk}^{\eta}
         \ell(dis_{\mathbf{M}}^2(\bm{x}_{i}^{t}, \bm{x}_{r}^{t})-dis_{\mathbf{M}}^2(\bm{x}_{i}^{t}, \bm{x}_{j}^{t})) \\&+ \frac{\lambda_2}{2} \left\| \mathbf{M}\right\|_{F}^{2}  \\
         &s.t. \quad \sum_{k =1}^K w_{ik} = 1; \ \forall{1 \leq i \leq N};
        \  \mathbf{M}\in \mathcal{S}_{d}^{+}.
     \end{split}
 \end{equation}
the square Frobenius norm of $\mathbf{M}$ is used as the regularization to mitigate the risk of over-fitting, the $\lambda_1 > 0$ and $\lambda_2 > 0$ are two hyperparameters. The sum of the product term $\sum_{k =1}^K w_{ik}^{\eta} w_{jk}^{\eta}$ is adopted as a weight imposed on the corresponding triplet loss. When the similar pair $(x_i^t, x_j^t)$ is close to a local area, the value of $w_{ik}^{\eta}$ and $w_{jk}^{\eta}$ would be large, the larger the weight term, the more attention will focus on this triplet constraint. which helps to focus on optimizing thous local triplet constraints.
\par
For practical optimization, the $\ell(\cdot)$ is instantiated as the smooth hinge loss \cite{43}:
\begin{equation} \label{smooth_fun}
    \ell(x) = \begin{cases}
    0 & \text{if} \quad  x \geq 1\\
    \frac{1}{2} - x & \text{if}\quad  x \leq 0 \\
    \frac{1}{2}(1-x)^2 & \text{otherwise},
    \end{cases}
\end{equation}
which serves as a surrogate for hinge loss. The smooth hinge loss maintains a large margin between the distance of the dissimilar pair to increase the separability.
\subsection{Optimization}
In this subsection, we adopt an alternating optimization strategy to solve the minimization problem in Eq. (\ref{new1}). More specifically, we alternatively update $\mathbf{M}$, $\mathcal{W}$ and $\mathcal{C}_{K}$ to optimize the objective function.\\
{\bfseries Initialization}: For the metric matrix $\mathbf{M}$, it can be initialized as the identity matrix. The weight set $\mathcal{W}$ and the local centers set $\mathcal{C}_{K}$ can be obtained by a Gaussian Mixture Modal (GMM) \cite{51} estimation. In particular, for a given data point $\bm{x}_{i}$, the $w_{ik}$ is the posteriori probability for $k$-th local component, and the $\bm{c}_{k} \in \mathbb{R}^{d}$ is the weighted mean of the instances belong to the $k$-th cluster.\\
{\bfseries Fixed $\mathbf{M}$ and $\mathcal{W}$ to solve $\mathcal{C}_K$}: We can observe that only the first term in Eq. (\ref{new1}) involves the $\mathcal{C}_K$, therefore, the minimization of this sub-problem can be expressed as:
\begin{equation}\label{pk}
\begin{split}
    \min \limits_{ \mathcal{C}_K}\mathcal{H}(\bm{c}_{k}) &= \sum_{i = 1}^N w_{ik}^{\eta}(\bm{x}_{i} - \bm{c}_{k} )^T\mathbf{M}(\bm{x}_{i} - \bm{c}_{k})\\
    &s.t. \quad \forall{1 \leq k \leq K}.
    \end{split}
\end{equation}
By computing the gradient of $\mathcal{H}(\cdot)$ with respect to $\bm{c}_{k}$, and set the gradient term to zero, we can get the close form solution of $\bm{c}_{k}$:
\begin{equation} \label{solve_Pk}
    \bm{c}_{k} = \frac{\sum_{i =1}^N w_{ik}^{\eta}\bm{x}_{i}}{\sum_{i =1}^Nw_{ik}^{\eta}}.
\end{equation}

{\bfseries Fixed $\mathbf{M}$ and $\mathcal{C}_K$ to solve $\mathcal{W}$}: For this sub-problem, we propose to use a alternatively way to solve the $\mathcal{W}$. More explicitly,
when we optimize the $\bm{w}_i$ ($\forall{i = 1,...,N})$, the other $\bm{w}_{j} (j \neq i)$ is fixed. Then, the sub optimization problem can be transformed as:
\begin{equation} \label{solve_W}
    \min \limits_{w_{ik}}\sum_{k = 1}^K w_{ik}^{\eta}F_{ik}, \quad
    s.t. \quad w_{ik} > 0, \sum_{k = 1}^K w_{ik} = 1.
\end{equation}
where $F_{ik} = \frac{1}{N}(\bm{x}_{i} - \bm{c}_{k})^T\mathbf{M}(\bm{x}_{i} - \bm{c}_{k}) + \frac{\lambda_1}{T}(\sum_{t = 1}^{T_1}w_{jk}^{\eta}\ell(\delta_{ijr}^t)+ \sum_{t = 1}^{T_2} w_{jk}^{\eta}\ell(\delta_{jir}^t))$ and $\delta_{ijr}^t = dis_{\mathbf{M}}^2(\bm{x}_{i}^{t}, \bm{x}_{r}^{t})-dis_{\mathbf{M}}^2(\bm{x}_{i}^{t}, \bm{x}_{j}^{t})$, it should be noticed that subscript in $\delta_{ijr}^t$ means the triplet constraints that contain the sample $\bm{x}_{i}$ as the first element in the triplet constraints. $T_{1}$ and $T_{2}$ are number of triplets that contain the sample $\bm{x}_{i}$ as first and second element in the given triplet constraints respectively.
\par
By using Lagrange multiplier $\xi$ to take the constraint $\sum_{k = 1}^K w_{ik} = 1$ into consideration, the Lagrange function could be written as follows:
\begin{equation}
   \mathcal{Q}(\bm{w}_{i}, \xi) = \sum_{k = 1}^K w_{ik}^{\eta}F_{ik} + \xi (\sum_{k = 1}^K w_{ik} - 1).
\end{equation}
\par
By setting the derivative of $\mathcal{Q}(\bm{w}_{i}, \xi)$ with respect to $w_{ik} (k = 1,\dots,K)$ and $\xi$ to zero, we have
\begin{equation}
    \begin{cases}
    \frac{\partial \mathcal{Q}}{\partial w_{i1}} = \eta w_{i1}^{\eta -1}F_{i1} - \xi = 0\\
    \vdots \\
    \frac{\partial \mathcal{Q}}{\partial w_{iK}} = \eta w_{iK}^{\eta -1}F_{iK} - \xi = 0\\
     \frac{\partial \mathcal{Q}}{\partial \xi} = \sum_{k =1}^K w_{ik} - 1= 0\\
    \end{cases}
\end{equation}
solving the above equation, we can get the closed form solution of $w_{ik} (k = 1, \cdots, K,i = 1, \cdots, N)$
\begin{equation}\label{solve_ck}
    w_{ik} = \frac{(1/F_{ik})^{\frac{1}{\eta -1}}}{\sum_{k = 1}^K (1/F_{ik})^{\frac{1}{\eta -1}}}.
\end{equation}
since $F_{ik} > 0$, then we have $w_{ik} > 0$. According to Eq. (\ref{solve_ck}), when $\eta \rightarrow +\infty$, $(w_{i1},\cdots, w_{iK})$ will be close to each other and $w_{ik} \rightarrow \frac{1}{K}$, which means that every locality make equal contribution to the sample $\bm{x}_{i}$; when
$\eta \rightarrow  1$, only the term $w_{ik}$ corresponding to the smallest $F_{ik}$ is close to $1$, the other elements in $(w_{i1},\cdots, w_{iK})$ will be close to zero.
\\
{\bfseries Fixed $\mathcal{C}_K$ and $\mathcal{W}$ to solve $\mathbf{M}$}: In Eq. (\ref{new1}), three parts are all involved with $\mathbf{M}$, we can derive the derivative of $\mathcal{L}$ with respect to $\mathbf{M}$ as
$\mathbf{G}_{M} = \mathbf{\nabla}_{1} + \lambda_{1} \mathbf{\nabla}_{2} + \lambda_{2} \mathbf{\nabla}_{3}$, where $\mathbf{\nabla}_{1}$, $\mathbf{\nabla}_{2}$ and $\mathbf{\nabla}_{3}$ are the corresponding three components in Eq. (\ref{new1}).
\begin{equation} \label{gradient}
    \begin{split}
        &\mathbf{\nabla}_1 = \frac{1}{NK}\sum_{i =1}^N \sum_{k=1}^K w_{ik}^{\eta}(\bm{x}_{i}- \bm{c}_{k})(\bm{x}_{i}- \bm{c}_{k})^T\\
        &\mathbf{\nabla}_2 = \frac{1}{KT}\sum_{k = 1}^K \sum_{t = 1}^T w_{ik}^{\eta}w_{jk}^{\eta}\mathcal{\ell}'(\delta_{ijr}^t)(\mathbf{A}_{ir}^{t} - \mathbf{A}_{ij}^{t})\\
        &\mathbf{\nabla}_3 = \mathbf{M}.
    \end{split}
\end{equation}
where $\ell'(x)$ is the derivative of smooth hinge loss function in Eq. (\ref{smooth_fun}), if $x \leq 0$, $\ell'(x) = -1$, and if $0 < x < 1$, $\ell'(x) = x -1$, the
$\mathbf{A}_{ir}^{t} = (\bm{x}_{i} - \bm{x}_{r})(\bm{x}_{i} - \bm{x}_{r})^T$ is the outer product of the difference of $\bm{x}_{i}$ and $\bm{x}_{r}$.
\par
To ensure the learned matrix is valid PSD matrix, most existing DML algorithms propose to solve this problem by the projected gradient \cite{1, 8, 29}, Let $\alpha$ be a proper step, we can update the $\mathbf{M}$ with $\mathbf{M}_{new} = \mathbf{M} - \alpha \mathbf{G}_{M}$ and then project the $\mathbf{M}_{new}$ into the PSD cone by using singular value decomposition. However, the sophisticated projected gradient process may encounter the numerical unstable problem \cite{31}. In light of the recent advance in Riemannian optimizaition technique, in this paper, we propose to optimize the $\mathbf{M}$ on the Riemannian manifold.
\par
\begin{figure}[htbp]
      \centering
      \includegraphics[width = 0.8\textwidth]{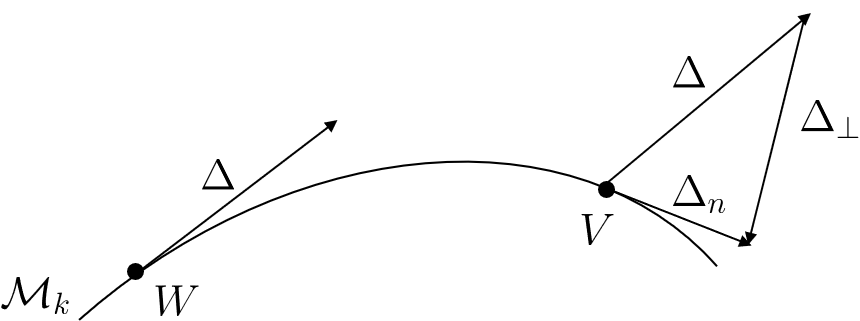}
      \caption{Parallel transport of a tangent vector $\mathbf{\Delta}$ from a point $\mathbf{W}$ to another point $\mathbf{V}$ on the manifold}.
      \label{para}
  \end{figure}

We propose to solve this sub-problem using Riemannian Conjugate Gradient Descent (RCGD) method. To perform RCGD optimization method on the manifold, we need to transport a tangent vector $\mathbf{\Delta}$ from one point to another point, unlike flat spaces, this procedure can't be done by simply translation. As illustrated in Fig. \ref{para}, transporting $\mathbf{\Delta}$ from $\mathbf{W}$ to $\mathbf{V}$ on the manifold $\mathcal{M}_k$, one needs to subtract the normal part of $\mathbf{\Delta}_{\bot}$ at $\mathbf{V}$, such a transfer on the tangent vector is called parallel transport \cite{38}, which is required by RCGD method to compute the new descent direction by combining the gradient direction at current with previous steps.\\
 we summarize the proposed RCGD in Algorithm\ref{rcgd}, where the $\tau(\cdot)$ means the parallel transport process, the whole proposed MDaML method can be found in Algorithm \ref{S2ml}. We implement the RCGD method with the manopt Riemannian optimization toolbox \cite{44}.
\begin{algorithm}
\caption{Riemannian Conjugate Gradient Descent (RCGD)}
\label{rcgd}
\begin{algorithmic}[1]
\REQUIRE
The initial metric matrix $\mathbf{M}_0 \in \mathcal{S}_{d}^{++}$, the parallel transport parameter $\beta$.\\
\ENSURE
Optimal solutions $\mathbf{M}^{*}$\\
\STATE Set $\mathbf{M} \leftarrow{\mathbf{M}_0}.$
\STATE Set $\mathbf{H}_{old} \leftarrow{\mathbf{0}}.$
\REPEAT
\STATE Compute the Euclidean derivative $\mathbf{G}_{M}$ by Eq.(\ref{gradient}).\\
\STATE Compute the Riemannian gradient $\hat{\mathbf{G}}_{M}$ by Eq.(\ref{riemann_gd}).\\
\STATE Find the searching direction by conjugate gradient descent:\\
$\mathbf{H}\leftarrow{-\hat{\mathbf{G}}_{M} + \beta\tau(\mathbf{H}_{old}, \mathbf{M}_{0}, \mathbf{M})}$.\\
Line search along the geodesic $\gamma(t)$ from $\gamma(0) = \mathbf{M}$ in the direction $\mathbf{H}$ to find $\mathbf{M}^{*} = \mathop{\arg\min}_{M}\mathcal{{\mathcal L}}(\mathbf{M})$ by Eq.(\ref{new1}).
\STATE Set $\mathbf{H}_{old} \leftarrow{\mathbf{H}}.$
\STATE Set $\mathbf{M}_{0} \leftarrow{\mathbf{M}}.$
\STATE Set $\mathbf{M} \leftarrow{\mathbf{M}^{*}}.$
\UNTIL{convergence}
\RETURN
\end{algorithmic}
\end{algorithm}

\begin{algorithm}
\caption{Multimodal-Aware Weakly Supervised Metric Learning with Self-weighting Triplet Loss}
\label{S2ml}
\begin{algorithmic}[1]
\REQUIRE
Training set $\mathbf{X}$, and triplet constraints set $\mathcal{T}= (\bm{x}_{i}^{t}, \bm{x}_{j}^{t}, \bm{x}_{r}^{t})_{t =1}^T$, number of locality $K$, trade-off parameters $\lambda_1$ and $\lambda_2$.
\ENSURE
Optimal solutions $\mathbf{M}^{*}$, the corresponding $\mathcal{W}^{*}$ and $\mathcal{C}_K^{*}$.\\
\STATE Initialize the $\mathcal{W}$ and $\mathcal{C}_{K}$ using the GMM algorithm, set $\mathbf{M} = \mathbf{I}$.
\REPEAT
\STATE Fixed $\mathbf{M}$ and $\mathcal{W}$ to solve $\bm{c}_{k} (k = 1, \cdots, K)$ by Eq. (\ref{solve_Pk}).\\
\STATE Fixed $\mathbf{M}$ and $\mathcal{C}_{K}$ to solve $w_{ik}(i = 1, \cdots, N; k = 1, \cdots, K)$ by Eq. (\ref{solve_ck}).\\
\STATE Fixed $\mathcal{C}_{K}$ and $\mathcal{W}$ to solve $\mathbf{M}$ by RCGD using Alogorithm \ref{rcgd}.
\UNTIL{convergence}\\
Set $\mathbf{M}^{*} \leftarrow{\mathcal{M}}$.\\
Set $\mathcal{C}_{K}^{*} \leftarrow{\mathcal{C}_{K}}$.\\
Set $\mathcal{W}^{*} \leftarrow{\mathcal{W}}$.\\
\RETURN
\end{algorithmic}
\end{algorithm}

\subsection{Complexity}
The time complexity of the proposed MDaML algorithm mainly lies in three parts: solving the three sub-problem in Algorithm \ref{S2ml}. The outer loop and inner loop of the proposed algorithm denote as $R_1$ and $R_2$ respectively, the whole time complexity depends on the following main step:\\
$\bullet$ For the fixed $\mathbf{M}$ and $\mathcal{W}$ to solve $\mathcal{C}_{K}$ using the Eq. (\ref{solve_Pk}) takes $\mathcal{O}(KNd)$.\\
$\bullet$ For the fixed $\mathbf{M}$ and $\mathcal{C}_{K}$ to solve $\mathcal{W}$ by the Eq. (\ref{solve_ck}), which involves the evaluation of $F_{ik}$. The time complexity of this term is $\mathcal{O}(NK(3d + 6Td))$. Where $T$ is the number of the triplet constraints.\\
$\bullet$ For the fixed $\mathcal{W}$ and $\mathcal{C}_{K}$ to solve $\mathbf{M}$. Computing the Euclidean gradient of $\mathcal{L}$ w.r.t $\mathbf{M}$ costs $\mathcal{O}(NKd^2 + 2KTd^2)$. Projecting the Euclidean gradient to tangent space takes $\mathcal{O}(2d^3)$. For the retraction part, the time complexity is $\mathcal{O}(3d^3)$, thus, the RCGD method takes $\mathcal{O}(5d^3 + NKd^2 + 2KTd^2)$.
\par
From the above analysis, we know that the total time complexity of the proposed algorithm is bounded by $\mathcal{O}(R_1(6NKTd + R_2(5d^3 + 2KTd^2)))$. In the real-world applications as $K$ is much smaller than $N$, the main cost of the MDaML is to update $\mathcal{M}$. In the experiment, we observed that the algorithm convergence very quickly, it usually took less than $5$ runs of outer iterations to convergence.
\par
Overall. Our proposed MDaML algorithm learn the $K$ locality centers and metric in a joint manner. During the training process, the proposed algorithm can adaptively learn the local neighbor for each mode. An alternating strategy is utilized to solve the algorithm, which is proven to be efficient by the experiment result in Section. \uppercase\expandafter{\romannumeral4}. \\

\section{Experiments}\label{Related Materials}
%% table of datasest and images
\begin{table}[htbp]
\centering
\caption{Summary statistics of the UCI datasets used in the experiment.}
\label{uci_data_sets}
\begin{tabular*}{0.6\textwidth}{@{\extracolsep{\fill}}lcccc}  % {lccc} left-l,right-r,center-c
\hline
Datasets & \# features & \#classes & \# examples  \\
\hline  % \hline
sonar & 60 &2 & 208\\
dna & 180 &3 & 2000  \\
balance & 4 & 3 & 625\\
autompg & 7 & 3 & 392\\
letter &16 &26 &5000\\
vehicle &18 & 4& 846\\
spambase & 57& 2& 4601\\
waveform& 21& 3&5000 \\
german & 24& 2& 1000 \\
segment &18 & 7&2310\\
\hline
\end{tabular*}
\end{table}

In order to verify the performance of MDaML, we conduct various experiments to evaluate the effectiveness of our proposed method in the context of $k$-NN classification setting. First, we run the classification experiment on 10 classification benchmark data sets. Then, we apply our method in three image datasets to validate the the effectiveness on the image datasets. Moreover, the empirical convergence of the proposed method is evaluated. Finally, we take some discussions on MDaML in the paper.

\subsection{Experiments setting}
To validate the performance of our method, we compare it with three weakly supervised DML approaches and four state-of-the-art supervised DML methods, including:\\
$\bullet$ LMNN is proposed in work \cite{5}, which aims to learn a metric such that the $k$ nearest neighbors of each training instance share the same label while samples from different classes are separated by a large distance.\\
$\bullet$ GMML is a weakly supervised DML method proposed in work \cite{6}, which aims to make the distance between examples of similar pairs as small as possible while keep the distance between examples of dissimilar pairs larger than a predefined threshold.\\
$\bullet$ KISSME assumes that examples from the similar pairs and dissimilar pairs are generated from two multivariate gaussian distribution, and the DML problem is converted into a maximizing a likelihood ratio \cite{8}.\\
$\bullet$ DML-eig formulates the DML as a convex optimization problem, whcih can be solved by repeated computing the largest the eigenvector of a matrix\cite{52}.\\
$\bullet$ DMLMJ is proposed in work \cite{15} to transforms the data into a new space, where the Jeffrey divergence of the two distribution "difference of positive space" and "difference of negative space" is maximized.\\
$\bullet$ LFDA is a localized variant of Fisher discriminant analysis (FDA), which takes the local structure of the data into account \cite{50}.\\
$\bullet$ DML-dc is a local DML method proposed in work \cite{53}, which intends to maximize the margin of nearest-neighbor classifier.\\
$\bullet$ EUCLID is the Euclidean distance metric, which is also recorded and serve as a baseline.
\par
Among all the methods, there are four weakly supervised algorithms, including: MDaML, GMML, KISSME, DML-eig, which need pairwise or triplet constraints as side information. To make a fair comparison, triplets constraints are generated by randomly choosing the $10$ same class samples and $10$ different classes samples for each training data, the pairwise set $\mathcal{S}$ and $\mathcal{D}$ are constructed by breaking down the triplet constraints $(\bm{x}_{i}, \bm{x}_{j}, \bm{x}_{r})$ as $(\bm{x}_{i}, \bm{x}_{j})$ and $(\bm{x}_{i}, \bm{x}_{r})$.
\par

\begin{figure*}[htbp]
\centering
\begin{minipage}{0.42\textwidth}
\centerline{\includegraphics[width=0.95\textwidth]{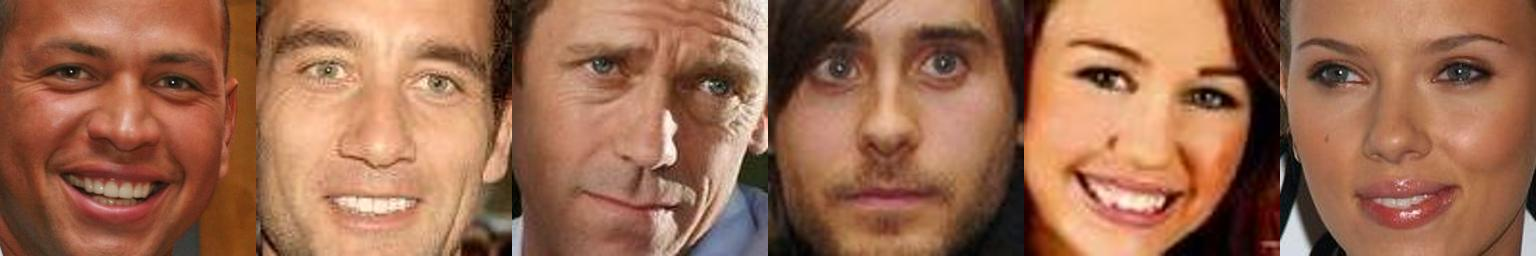}}
\label{sample.1}
\centerline{(a) pubfig}
\vfill
\centerline{\includegraphics[width=0.95\textwidth]{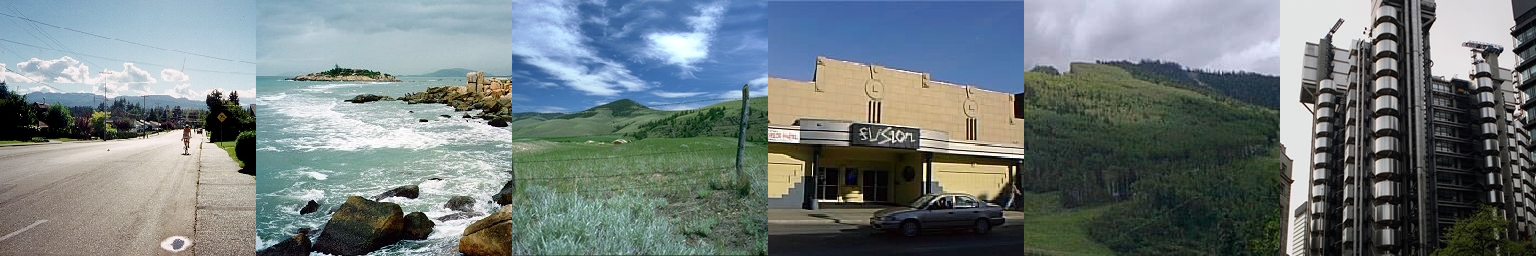}}
\label{sample.2}
\centerline{(b) osr}
\end{minipage}
\hfill
\begin{minipage}{0.56\linewidth}
\centerline{\includegraphics[width=0.9\textwidth]{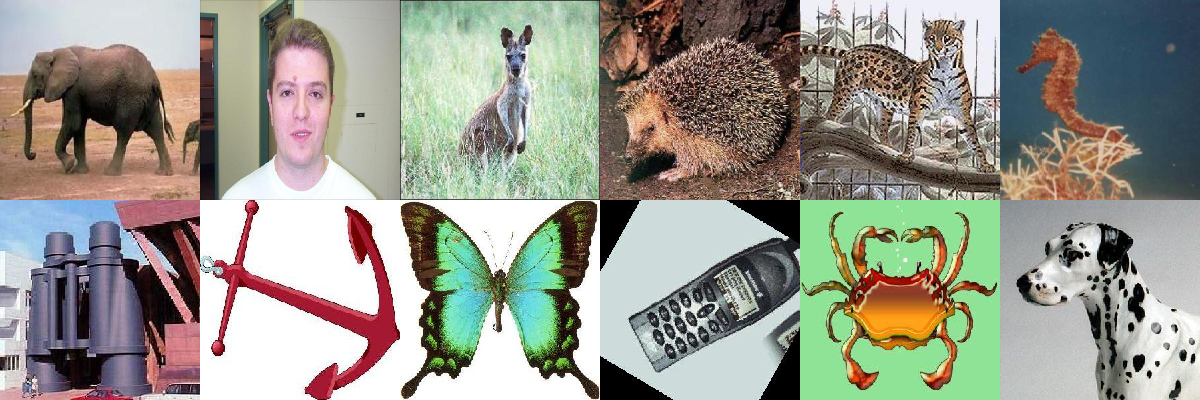}}
\label{sample.3}
\centerline{(c) caltech}
\end{minipage}

\caption{Sample images from three image datasets used in the experiment.}
 \label{figure_sample}
\end{figure*}

\subsection{Experiments on benchmark datasets}
All the methods are evaluated on the 10 real-world datasets from the UCI repository \cite{54}. The detailed information of each dataset including number of instances, the number of features, and classes counts can be found in Table \ref{uci_data_sets}. The test accuracy on each dataset is repeated 10 trials, at each trial, we randomly partition each dataset into two parts with partition $\{70\%, 30\%\}$, the first part is for training and the second is for testing. Parameters of all methods are tunes in the first trial on the training data and fixed for all the remaining trials. For MDaML, there are four hyper-parameters, based on the empirical experiments, the $\eta$ is set to $3$, and $\lambda_2$ is set to $0.0001$, which yields the best result on most datasets. The number of locality centers is tuned form the set $\{2, 4, 6, 8, 10, 15, 20, 25, 30, 40, 50\}$, and the $\lambda_1$ is choosing form $\{0.1, 1, 10, 100, 1000\}$. All datasets are normalized to have zeros mean and unit variance over the training data. At the test stage, $3$NN classification accuracy is used for assessment.
\par
The test accuracy of each method is measured by counting the number of the correctly assigned test sample points and then dividing the result by the number of all the points in the test dataset, which can be calculated as follow:
\begin{equation}
    \text{Acc} = \frac{\sum_{i = 1}^{n_e}(\delta(y_i = \hat{y_i}))}{n_e}
\end{equation}
where $n_e$ is the number of test data points, $y_i$ is the true class label of the data point $x_i$, $\hat{y_i}$ is the predicted label. $\delta(x)$ returns $1$ if $x$ is true, otherwise returns 0.

\begin{sidewaystable}
\centering
\caption{\normalsize Classification accuracies(test accuracies, mean $\pm$std.) based on 3NN. Every row represents the classification performance of different methods on a particular data set. The best performance on each dataset is in bold.}
\label{uci_res}
\begin{tabular}{cccccccccccccc}\hline
     & MDaML  & EUCLID & DMLMJ & GMML & KISSME & LMNN   & DML-eig  & LFDA & DML-dc \\
\hline  % \hline
autompg &\textbf{ 77.73$\pm$1.76}  &70.76$\pm$1.24 &73.94$\pm$2.30 &72.61$\pm$3.55 &73.95$\pm$2.24 &72.61$\pm$2.94 &72.18$\pm$3.05 &75.13$\pm$2.60 &74.03$\pm$2.76\\
balance &90.79$\pm$3.19       &82.33$\pm$2.65     &88.20$\pm$1.93     &87.14$\pm$2.03        &89.31$\pm$1.81         &86.03$\pm$2.41       &89.47$\pm$2.80      &\textbf{92.12$\pm$2.11}        &88.57$\pm$2.25        \\
dna &93.74$\pm$0.96        &72.38$\pm$1.94      &81.06$\pm$1.78      &82.06$\pm$1.63        &89.14$\pm$0.76         &\textbf{94.30$\pm$0.91}       &92.36$\pm$1.31      &93.74$\pm$0.85        &78.25$\pm$0.73        \\
german &\textbf{73.52$\pm$1.81}        &69.80$\pm$1.51      &70.03$\pm$1.83      &69.34$\pm$1.81        &70.63$\pm$1.62         &70.30$\pm$1.80       &72.46$\pm$2.60      &72.46$\pm$1.91        &70.03$\pm$2.85        \\
letter &92.83$\pm$0.53        &84.20$\pm$0.69      &90.36$\pm$0.80      &88.73$\pm$0.83        &92.16$\pm$0.78         &87.02$\pm$0.64       &90.00$\pm$0.97      &89.42$\pm$0.76        &\textbf{92.99$\pm$0.66}        \\
spambase &\textbf{92.12$\pm$0.52}        &84.53$\pm$0.89      &84.03$\pm$0.56      &85.58$\pm$0.87        &85.37$\pm$1.30         &86.05$\pm$1.67       &92.32$\pm$0.41      &73.82$\pm$7.11        &91.39$\pm$0.50        \\
sonar &83.13$\pm$3.36        &80.94$\pm$2.83      &80.00$\pm$4.09      &80.00$\pm$2.83        &79.06$\pm$5.67         &82.19$\pm$3.31       &82.03$\pm$2.88      &78.44$\pm$3.52         &\textbf{84.84$\pm$2.66}        \\
segment &\textbf{96.83$\pm$0.93}        &93.24$\pm$0.46      &94.54$\pm$0.86      &95.23$\pm$0.61        &94.84$\pm$1.41         &94.80$\pm$0.59       &94.57$\pm$1.06      &94.23$\pm$0.90        &96.26$\pm$0.76        \\
vehicle &\textbf{79.88$\pm$1.73}        &70.16$\pm$2.60      &77.07$\pm$2.00      &78.91$\pm$2.44        &78.79$\pm$1.67         &78.09$\pm$2.21       &69.26$\pm$2.59      &79.61$\pm$1.42        &74.30$\pm$2.78        \\
waveform &\textbf{84.04$\pm$0.81}        &79.56$\pm$0.63      &79.47$\pm$0.92      &83.06$\pm$0.30        &82.67$\pm$0.72         &80.75$\pm$1.01       &83.66$\pm$0.91      &82.45$\pm$0.73        &81.68$\pm$1.02        \\\hline
\end{tabular}
\end{sidewaystable}

\begin{figure*}[htbp]
  %\centering
  \subfigure[spambase]{
  \label{function_value.1}
 \includegraphics[width = 0.32\textwidth]{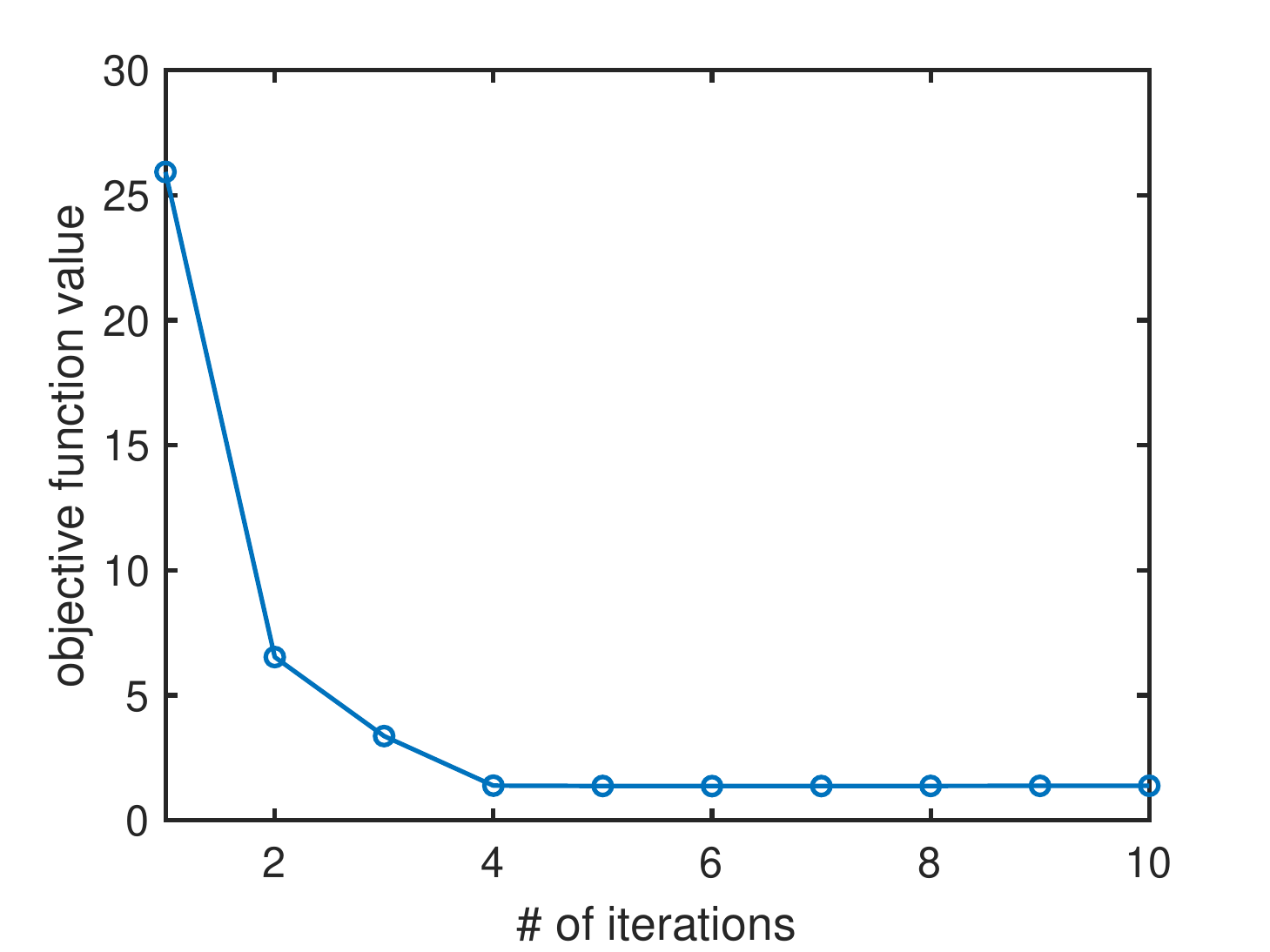}}
  \subfigure[letter]{
\label{function_value.2}
 \includegraphics[width = 0.32\textwidth]{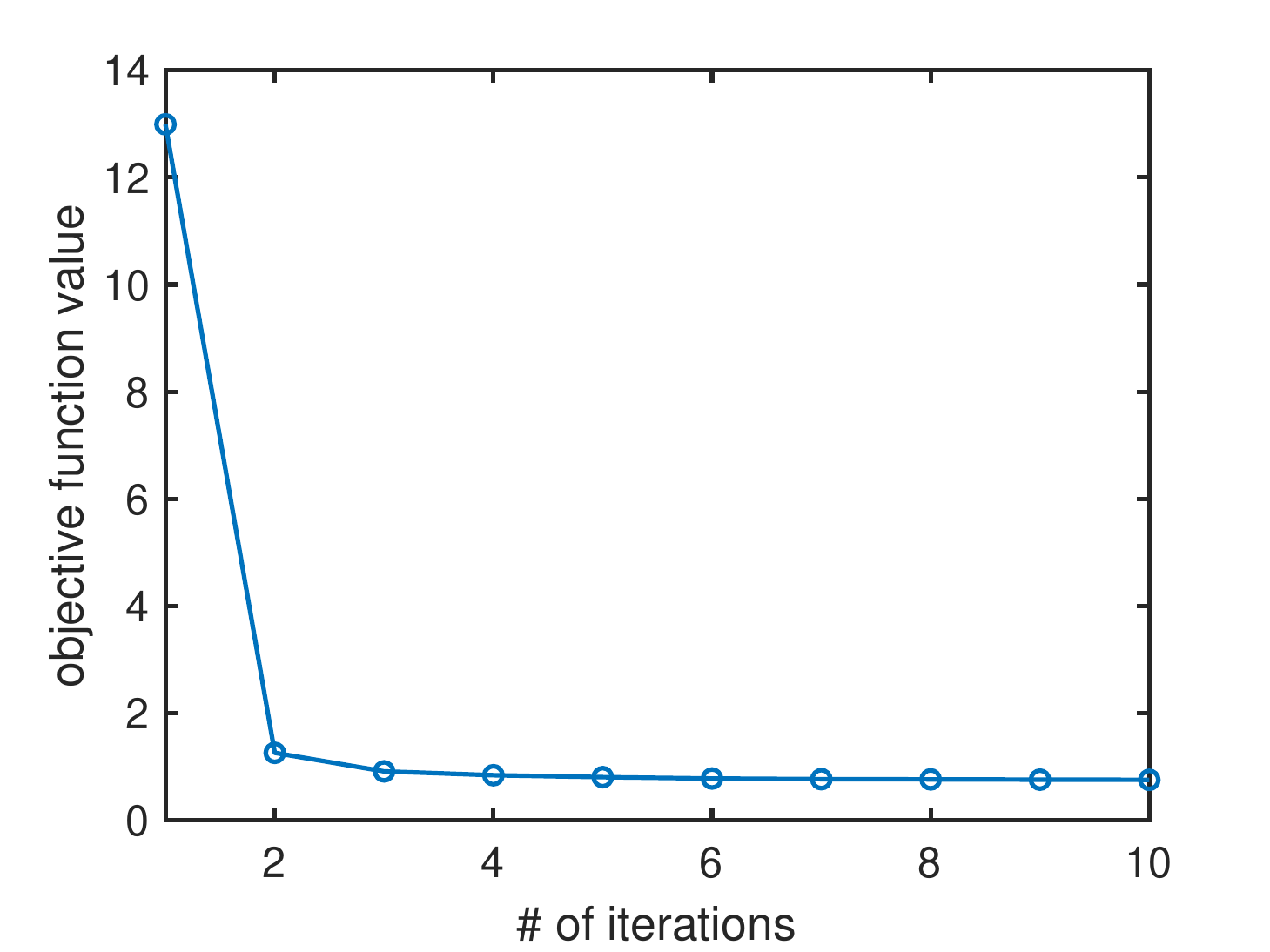}}
  \subfigure[dna]{
    \label{function_value.3}
 \includegraphics[width = 0.32\textwidth]{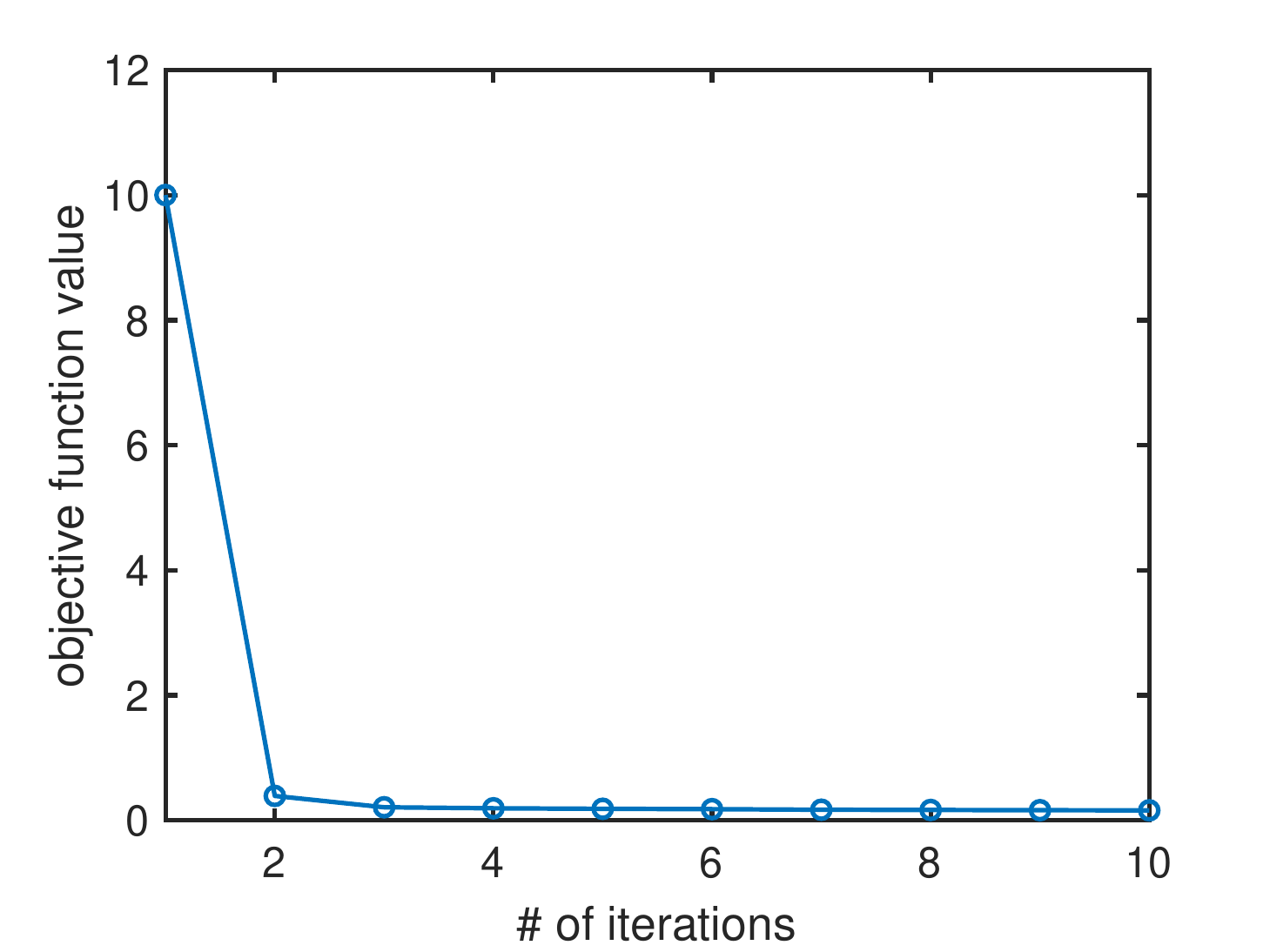}}
  \caption{The convergence behavior of proposed method in optimization on three datasets: (a) spambase; (b) letter (c) dna.}
  \label{function_value}
\end{figure*}
The average classification accuracy and standard derivations are recorded in Table \ref{uci_res}, the best performance on each dataset is in bold. From the table, it can be found that DML methods can improve the nearest neighbor classification performance compared with EUCLID in most cases. Compared with the three weakly supervised DML methods (GMML, KISSME, DML-eig), our method gains a significant improvement on most datasets. Since our proposed method could wisely identify those local triplet constraints, and optimize the local compactness and separability, which is more suitable for local classifiers like $k$-NN. Compared with supervised DML methods, our method achieves a competitive result. As in autompg, DML-dc and LMNN can significantly improve the classification accuracy than the EUCLID metric, while our proposed method can make further improvement. MDaML achieves the best performance on 6/10 datasets, which confirms the effectiveness of MDaML. Nevertheless, the performance of DMLMJ is slightly worse than that of the others, this may be explained by the fact that the strong assumption of the data distribution does not hold for the real-world application. On the contrary, the DML-dc also obtains promising results on most datasets, which may suggest the importance of learning a suitable nearest neighbor for DML.
\par
MDaML is solved via an alternative optimization strategy, the objective function value in Eq. \ref{new1} will decline once $\mathcal{C}_{K}$, $\mathcal{W}$ or $\mathbf{M}$ is updated, however, the loss objective function is non-negative, in the end, the proposed algorithm should be converged. To analyze the empirical convergence performance of MDaML, we plot the objective function value versus the number of outer iterations in Figure. \ref{function_value}. From the figure, it is observed that our method takes less than five runs to converge, on most datasets. We also record the training time of each method on the datasets in Table \ref{training_time}. It should be noticed that the training time doesn't include the time for tuning hyper-parameters. All the algorithms are conducted on the Matlab 2018b platform on a PC with a Core i7-8700 processor and 16 GB physical memory. GMML, KISSE, and LFDA have closed-form solutions, therefore, these methods have a relatively less training time than that of those iterative methods. It can be found that the DML-dc is significantly slower than the other algorithms, which is probably due to the requirement of computing the nearest neighbor and nearest imposter in each iteration. While our method is slightly slower than DML-eig in most cases and can be trained in a reasonable time.

\begin{table*}[htbp]
\centering
\caption{Training time (in seconds) of the MDaML and other metric learning methods.}
\label{training_time}
\begin{tabular*}{0.99\textwidth}{@{\extracolsep{\fill}}lcccccccc}  % {lccc} left-l,right-r,center-c
\hline
      &MDaML  & GMML & KISSME & LMNN   & DML-eig  & LFDA & DML-dc \\
\hline  % \hline
autompg &2.00   &0.005  &\textbf{0.003}  &0.261 &0.383&0.009 &5.36\\
balance &2.40   &\textbf{0.0017}  &0.0026  &0.081 &1.249&0.0038 &12.27 \\
dna     &43.4   &0.215  &0.674  &2.43 &39.0&\textbf{0.0600} & 2180\\
german &2.90   &0.023  &\textbf{0.016}  &0.29 &2.93 &\textbf{0.016} & 43.60\\
letter &102.53  &\textbf{0.0065}  &0.0216  &3.859 &25.781&0.0302 &1016.2\\
spambase &34.77   &\textbf{0.0124}  &0.128  &0.535 &17.93&0.482 &1488.7\\
sonar &1.21   &\textbf{0.0022}  &0.010  &0.185 &5.46&\textbf{0.0022} &3.1535\\
segment &18.85   &\textbf{0.002}  &0.059  &1.622 &7.11&0.013 &211.64\\
vehicle &3.682   &\textbf{0.0125}  &0.0141  &0.8036 &4.580&0.0141 &27.257\\
waveform3 &45.078  &\textbf{0.0019}  &0.0297  &11.033 &9.876&0.185 &1004.3\\
\hline
\end{tabular*}

\end{table*}

\subsection{Experiments on image recognition}

\begin{figure*}[htbp]
  %\centering
  \subfigure[pubfig]{
  \label{clssification.1}
 \includegraphics[width = 0.32\textwidth]{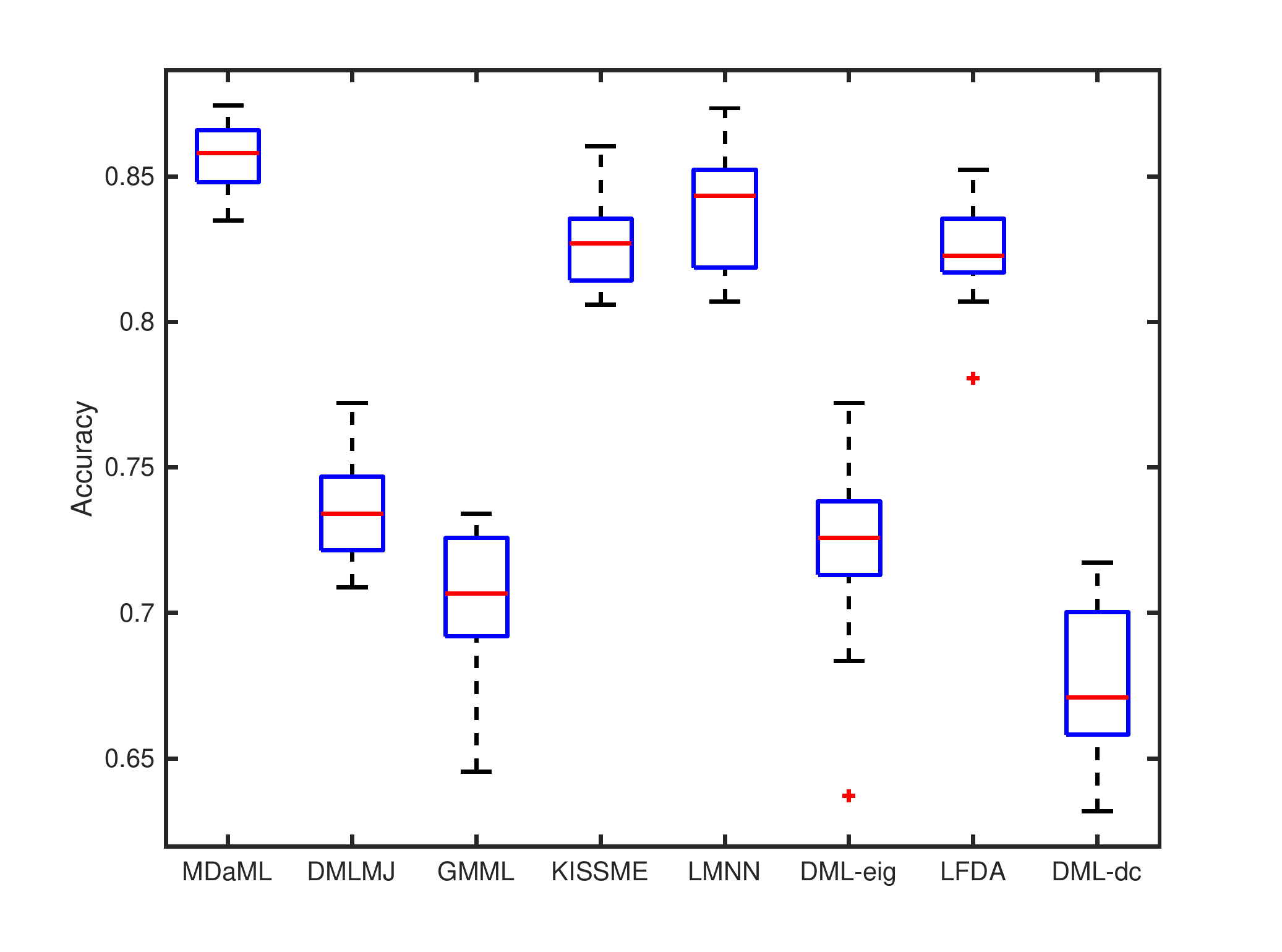}}
  \subfigure[osr]{
\label{clssification.2}
 \includegraphics[width = 0.32\textwidth]{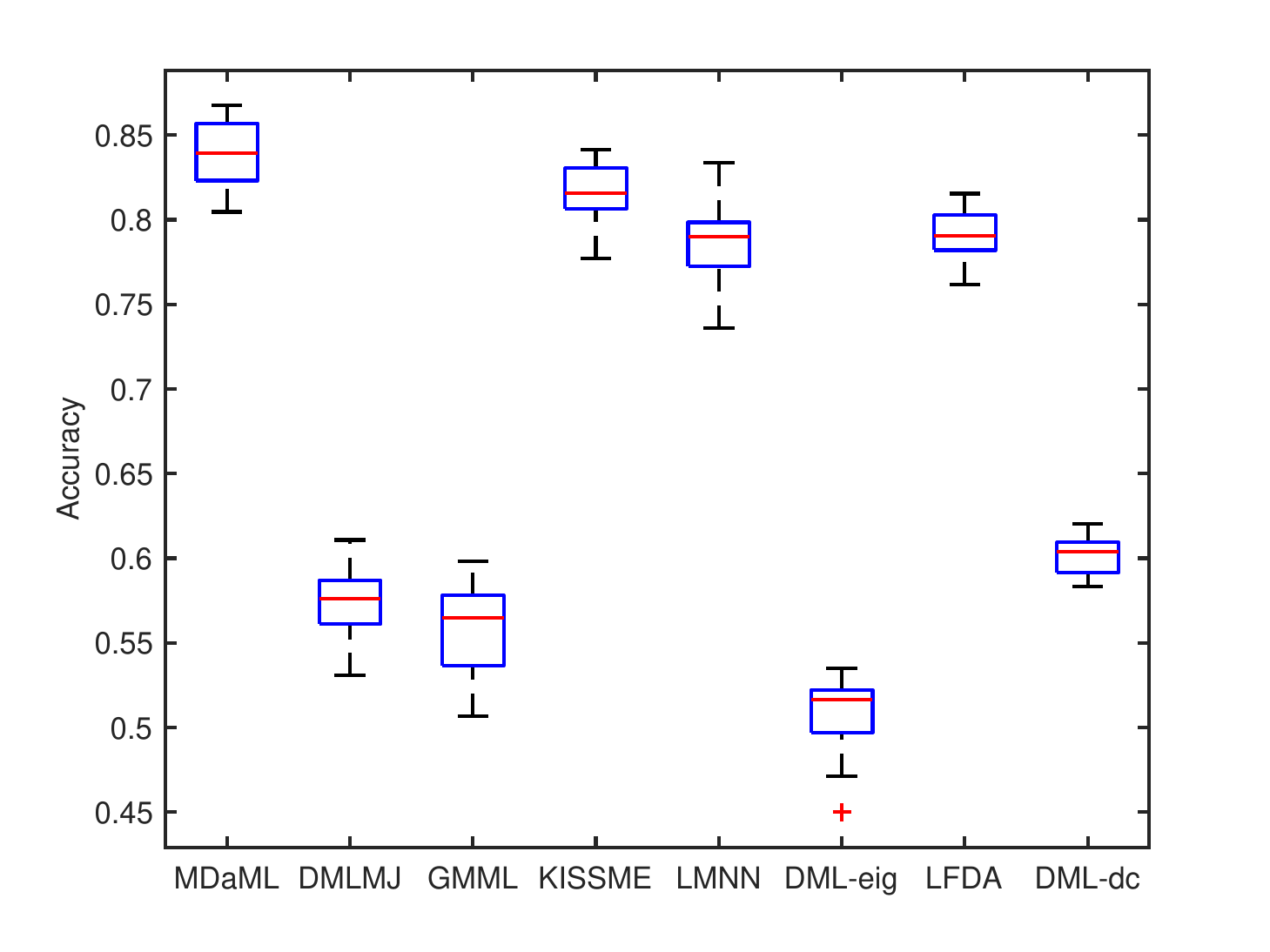}}
  \subfigure[caltech]{
    \label{clssification.3}
 \includegraphics[width = 0.32\textwidth]{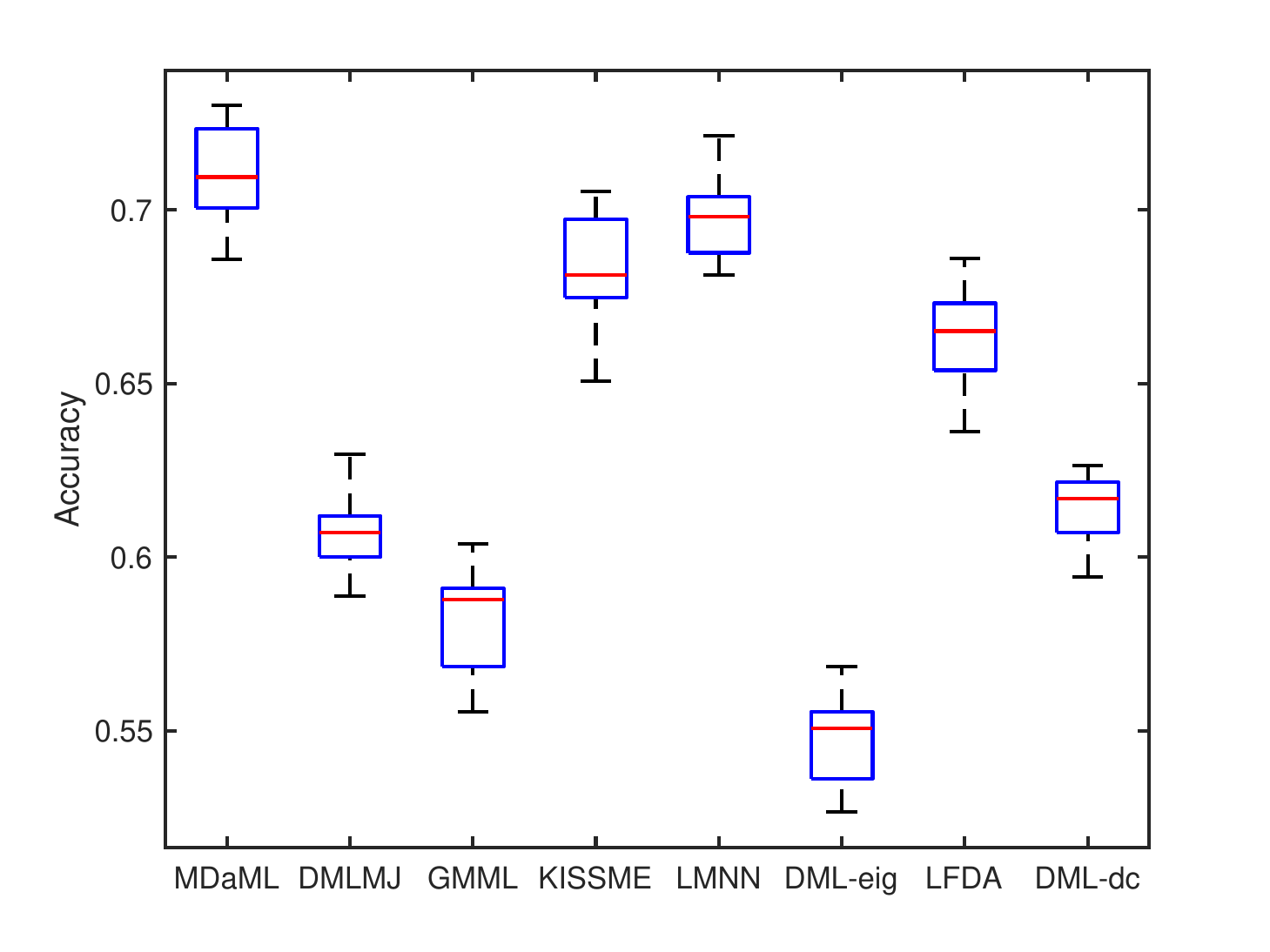}}
  \caption{Classification result on three datasets: (a) pubfig; (b) osr; (c) caltech.}
  \label{image_classification_res}
\end{figure*}

\begin{figure}[htbp]
  \centering
  \includegraphics[width = 0.8\textwidth]{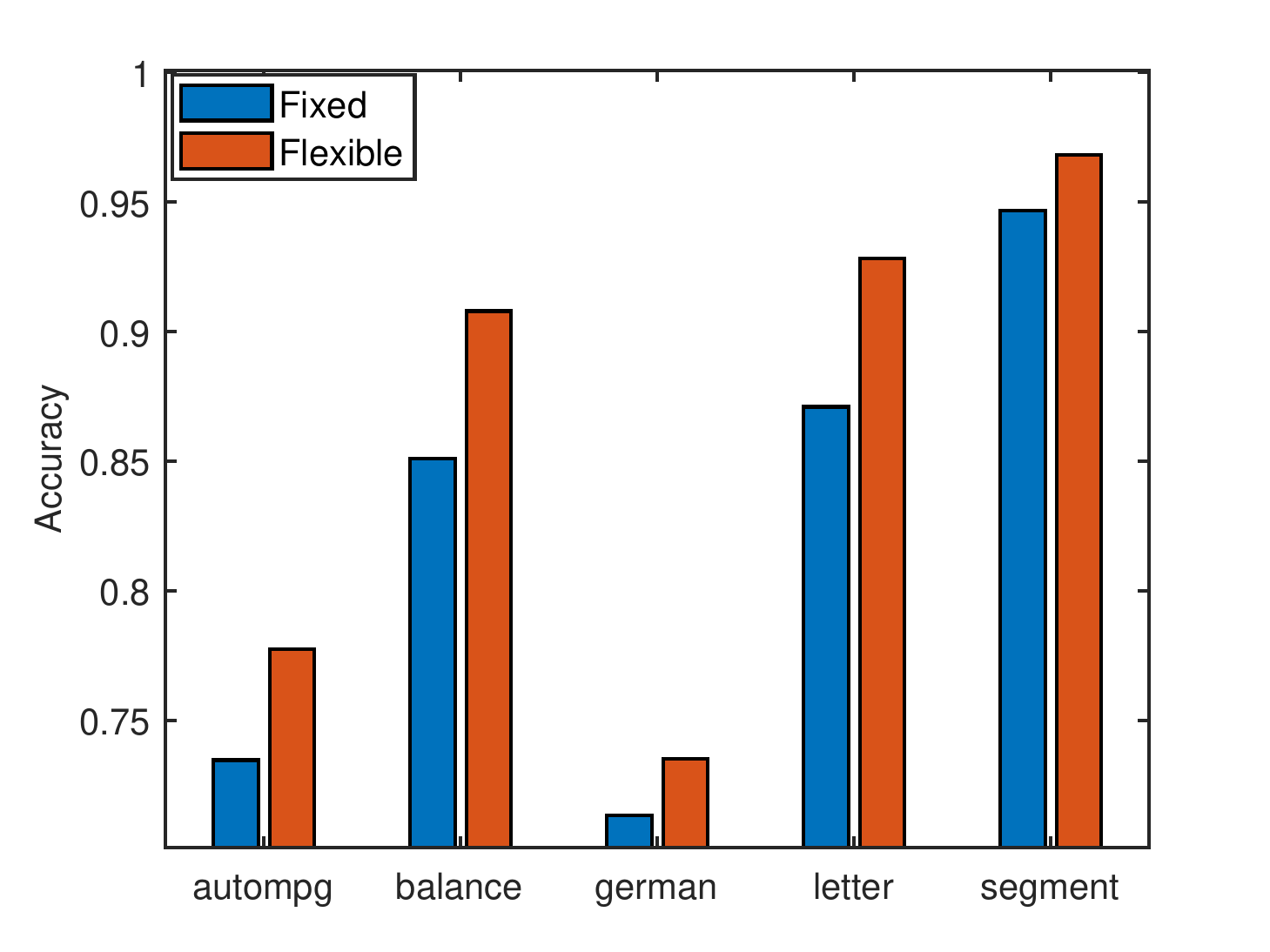}
  \caption{The influence of the self-weighting for the proposed method on the five benchmark datasets. "Fixed" means that all the weights of triplet constraints are fixed as 1, and "Flexible" means that all the weights are learned automatically.}
  \label{fixed_weight}
\end{figure}

\begin{figure*}[htbp]
  \centering
  \subfigure[]{
  \label{K}
 \includegraphics[width=5.5cm, height=3.8cm]{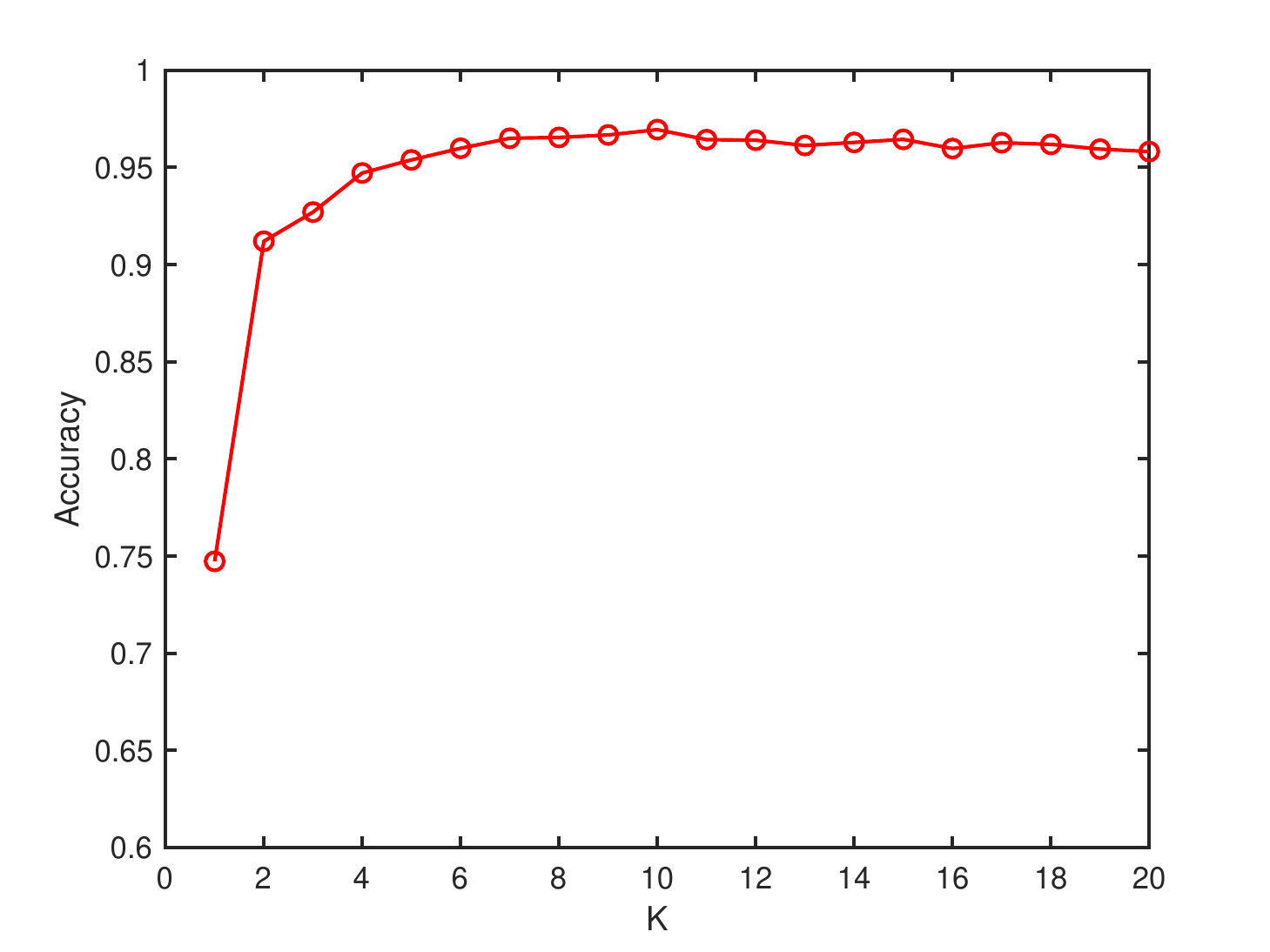}}
  \subfigure[]{
\label{lambda_1}
 \includegraphics[width=5.5cm, height=3.8cm]{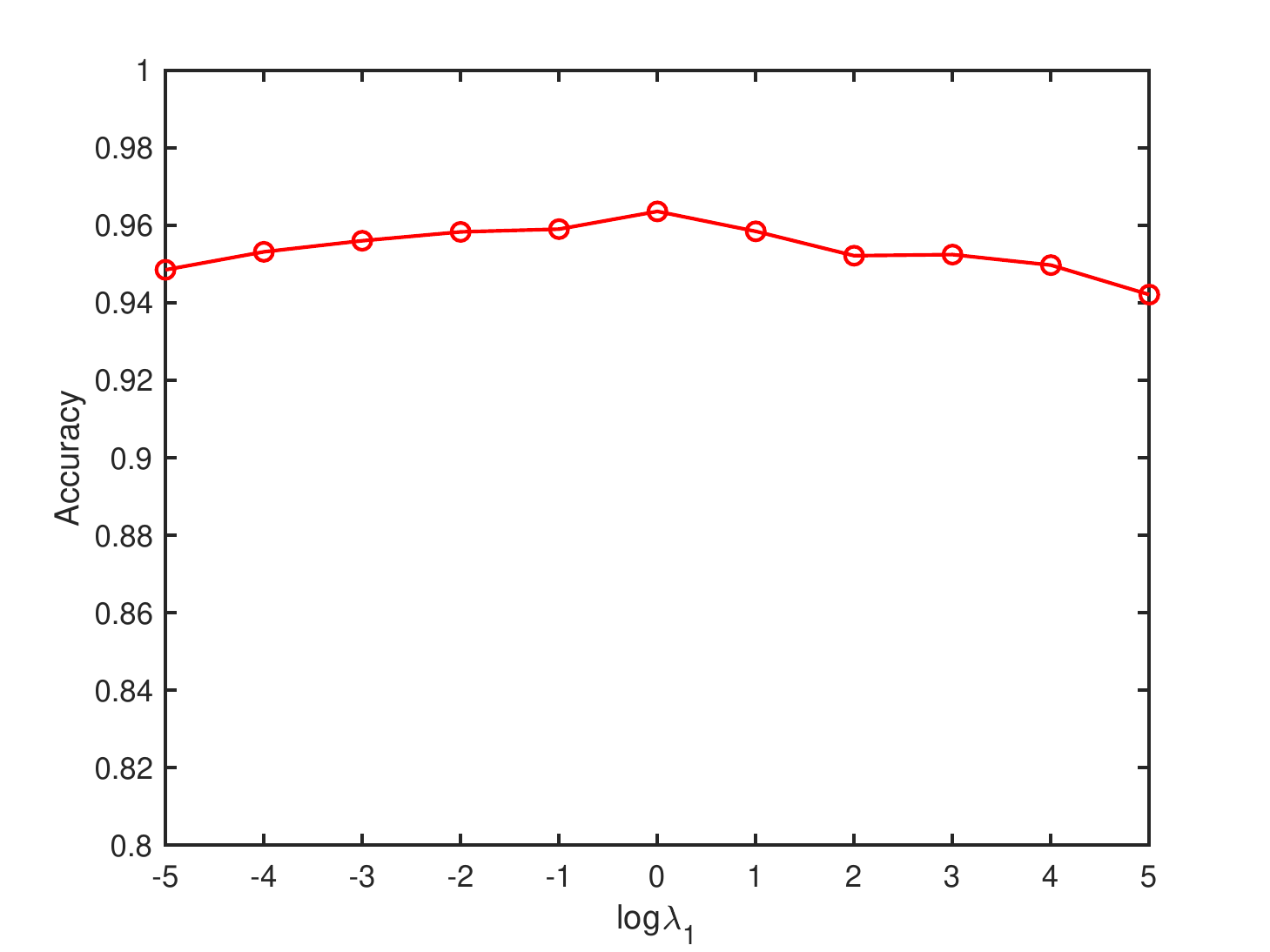}}
  \subfigure[]{
    \label{eta}
 \includegraphics[width=5.5cm, height=3.8cm]{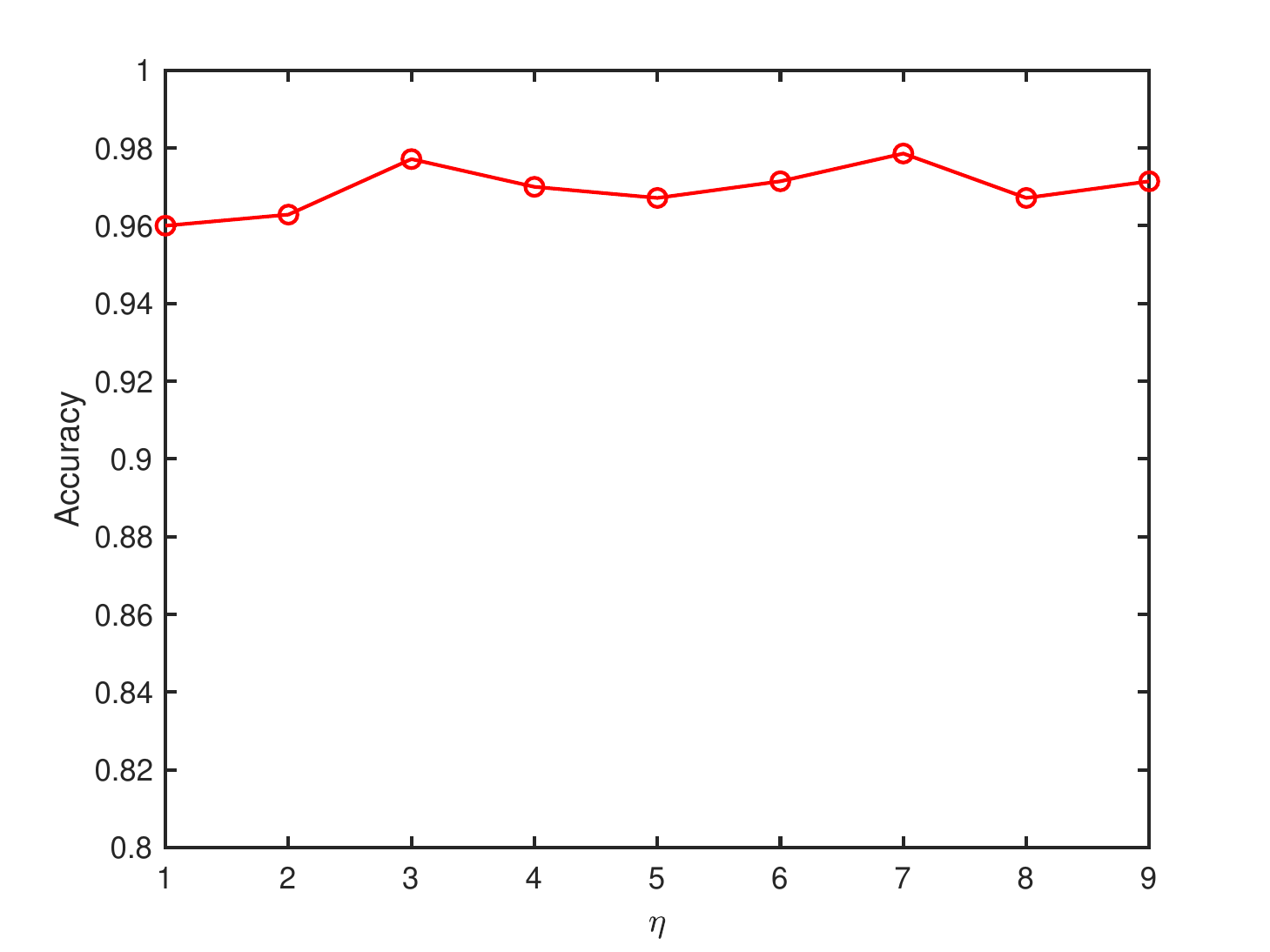}}
   \subfigure[]{
    \label{lambda_2}
 \includegraphics[width=5.5cm, height=3.8cm]{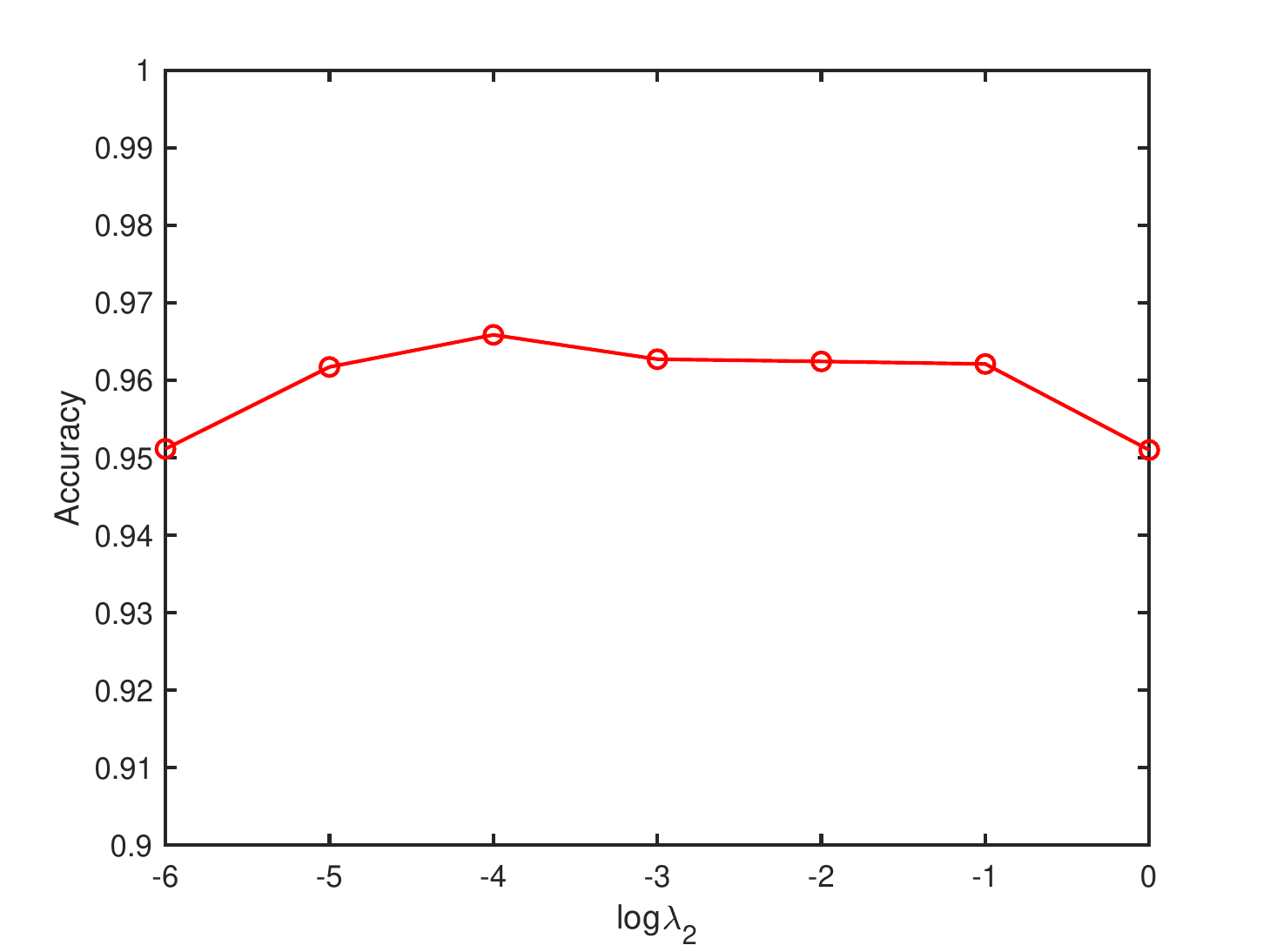}}
  \caption{Parameter sensitivity of the proposed method in the segment dataset: (a) varying $K$ while setting $\lambda_1$ = 1, $\eta$ = 3, $\lambda_2 = 10^{-3}$; (b) varying $\lambda_1$ while setting $K = 10$, $\eta$ = 3, $\lambda_2 = 10^{-3}$; (c) varying $\eta$ while setting $K = 10$,  $\lambda_1 = 1$, $\lambda_2 = 10^{-3}$; (d) varying $\lambda_2$ while setting $K = 10$,  $\eta = 3 $, $\lambda_1 = 1$.}
  \label{parameter_sensity}
\end{figure*}

 In this subsection, we focus on the image recognition task where the goal is to assign an image to a predefined class in the training data.
There are three image datasets used in the experiment, "pubfig" is the Public Figures Face Database \cite{11}, which is taken in completely uncontrolled situations with non-cooperative subjects. We adopt the same experiment setting as \cite{10}: a subset of face images of pubfig datasets is used, and there are 772 images from 8 categories. We extract 512-dimensional gist features and 45-dimensional Lab color histogram features, and concatenate them as the image features.
"OSR" is the Outdoor Scene Recognition Dataset, which consists of $2688$ images from $8$ scene categories. A 512-dimensional gist descriptor is used as image features.
"caltech 20" is a subset of the dataset of the original Caltech datasets\cite{13}. There are 20 classes of total 1536 images, we extract the 512-dimensional gist feature as image features. The detailed information of each datasets including the number of instances, the number of features, and classes counts is summarized in Table \ref{image_dataset}. Figure. \ref{figure_sample} shows some image samples from these three datasets.
For all the datasets, we use principal component analysis as a pre-processing step to reduce the dimensionality of the feature vectors to 100.
\begin{table}
\caption{Summary statistics of the image datasets used in the classification experiment.}
\label{image_dataset}
\centering
\begin{tabular*}{0.6\textwidth}{@{\extracolsep{\fill}}lcccc}  % {lccc} left-l,right-r,center-c
\hline
Datasets & \# features & \#classes & \# examples \\
\hline  % \hline
Pubfig & 542 & 8 & 772\\
OSR  & 512 &8 & 2688 \\
Caltech 20 & 512 & 20 & 1536\\
\hline
\end{tabular*}
\end{table}
We randomly partition the datasets into two part with partition $\{60\%, 40\%\}$, the first part is used for training and the second part is for evaluating. 3NN classification accuracy on each dataset is recorded and this procedure on each dataset is repeated 10 times.
\par
Figure. \ref{image_classification_res} shows the recognition results of the competing methods. As we can see from the results, our proposed method obtains the best results on the three image datasets, which indicates the effectiveness of our proposed method in high-dimensional datasets. It can be found that the performance of DML-eig and GMML have a severe drop compared with LMNN, which may be explained by the existence of multimodal distribution classes in the datasets. On caltech, MDaML outperforms GMML and DML-eig $14\%$ and $17\%$ on average, respectively. The performance of Dml-dc is significantly worse than that of the LMNN, one possible explanation is that DML-dc only pushes the nearest imposter away from the nearest neighbor and not enforces the same class samples aggregate close to each other.
\par
Overall, in the weakly supervised setting, enforcing the metric matrix to satisfy all the side information is not appropriate, and sometimes this goal would degenerate the performance of DML. It is readily seen that optimizing the local compactness and separability during the training process could be more effective for the weakly supervised DML setting.

\subsection{Discussion of self-weighting scheme}
Our proposed method adopts the self-weighting triplet loss to identify the appropriate local triplet constraints with the given side information. It is essential to investigate the influence of the self-weighting scale factor $w_{ik}^\eta$ on the performance of the proposed method.
\par
To show the importance of the self-weighting scale parameters. We fix all the $w_{ik}^\eta = 1, 1 \leq i \leq N, 1 \leq k \leq K$
and show the performance of MDaML on the benchmark datasets in Figure. \ref{fixed_weight}. For a fair comparison, all the experiment settings are the same as MDaML and the best parameters are tuned on the first trial.
\par
Through the Figure. \ref{fixed_weight}, it can be observed that the self-weighted strategy of the proposed method can achieve higher accuracy than that obtained by simply fixing all the weights of $w_{ik}^\eta$ to 1, which proves the effectiveness of the self-weighting schema for MDaML.

\subsection{Sensitivity analysis of the hyper-parameters}
There are four hyper-parameters in our method, namely: $K$, $\lambda_1$, $\lambda_2$, $\eta$, in this section, we discuss the effect of relevant parameters setting.
\par
On the parameter $K$: We introduce in Eq. \ref{cluster} a scale parameter $K$ that is expected to be the number of locality cluster centers of the training data. We compare the 3NN classification accuracy in the segment dataset with varying $K$ while setting the other parameters fixed. From the Figure. \ref{K}, we can see that the performance improves when the number of clusters increases, the best performance is reached when $K = 10$, it is curious to see that the number of classes of the segment is $7$, which may suggest that allocating more locality centers could help increase the classification accuracy to some extent. However, the performance tends to be stable when more $K$ are set.
\par
On the parameter $\lambda_1$, $\eta$, and $\lambda_2$: We use the same strategy as above, setting the other three parameters fixed and plot the classification accuracy with the varied parameter in Fig. \ref{eta}, Fig. \ref{lambda_1}, and \ref{lambda_2}. It can be found that these three hyper-parameters are not sensitive to $\lambda_1$, $\eta$, and $\lambda_2$ in the selected interval. For example, when $\eta > 2$, the classification accuracy only change within $2\%$ when we change the parameter $\eta$ from 3 to 10.
\par
Overall, the $\eta$, $\lambda_1$, and$\lambda_2$ are less sensitive than that of $K$. Consequently, we only tune the $K$ and $\lambda_1$, and empirically set $\eta$ and $\lambda_2$ as 3, and $10^{-3}$ respectively.

\section{Conclusion}
This paper focuses on how to obtain an appropriate distance metric from weakly supervised data. The key to resolve this issue is to fully exploit the underlying data distribution to identify those local side information.
MDaML adopts the clustering like term to partition the data space into several clusters based on the classes or the modes, combining self-weighted triplet loss with the clustering term, to further enhance the local compactness and separability of the algorithm. An iterative optimization strategy is proposed to solve MDaML. The experimental results demonstrate that the proposed method can achieve superior performance over other existing DML methods.
\par
In future, we will attempt to utilize other types of weight combination of $w_{ik}$, and apply our method in the pairwise or quadruplet weakly supervised datasets.

\begin{acknowledgements}
The authors would like to thank the anonymous reviewers for their insightful comments and the suggestions to significantly improve the quality of this paper. This work was supported by National Natural Science Foundation of PR China(61972064) and LiaoNing Revitalization Talents Program(XLYC1806006).
\end{acknowledgements}

\bibliographystyle{spphys}
\bibliography{paper}

\end{document}